\documentclass[runningheads]{llncs}

\usepackage{accv}

\usepackage[dvipsnames]{xcolor}
\usepackage{algorithm}
\usepackage{algpseudocode}
\usepackage{amsmath}
\usepackage{array}
\usepackage{graphicx}
\usepackage{subcaption}
\usepackage{tabularx,colortbl}
\usepackage{xcolor}
\usepackage{wrapfig}
\usepackage[final]{pdfpages}

\newcommand{\red}[1]{{\color{red}#1}}

\newcommand{\mypara}[1]{\vspace{2pt}\noindent{\bf{#1}}}
\newcommand{\cbd}{PopCharacters\xspace}
\newcommand{\pmx}{PopManga-X\xspace}

\definecolor{grey_cell}{RGB}{230,231,231}

\newlength\savewidth\newcommand\shline{\noalign{\global\savewidth\arrayrulewidth
  \global\arrayrulewidth 1pt}\hline\noalign{\global\arrayrulewidth\savewidth}}

\usepackage{etoolbox}

\newcommand{\uniformtablesize}{\fontsize{7}{9}\selectfont}

\AtBeginEnvironment{tabular}{\uniformtablesize}

\usepackage{accvabbrv}

\usepackage{graphicx}
\usepackage{booktabs}

\usepackage[accsupp]{axessibility}  

\usepackage{hyperref}

\usepackage{orcidlink}

\begin{document}

\title{Tails Tell Tales: Chapter-Wide Manga Transcriptions with Character Names} 

\titlerunning{Tails Tell Tails}

\author{Ragav Sachdeva \and
Gyungin Shin\thanks{Core contribution} \and
Andrew Zisserman
}

\authorrunning{Sachdeva et al.}

\institute{Visual Geometry Group, Dept. of Engineering Science, University of Oxford}

\maketitle

\begin{abstract}
\vspace{-1.5em}
Enabling engagement of manga by visually impaired individuals presents a significant challenge due to its inherently visual nature. With the goal of fostering accessibility, this paper aims to generate a dialogue transcript of a complete manga chapter, entirely automatically, with a particular emphasis on ensuring narrative consistency. This entails identifying (i) {\em what} is being said, i.e., detecting the texts on each page and classifying them into essential vs non-essential, and (ii) {\em who} is saying it, i.e., attributing each dialogue to its speaker, while ensuring the same characters are named consistently throughout the chapter. 

\quad To this end, we introduce: (i)  Magiv2, a model that is capable of generating high-quality chapter-wide manga transcripts with named characters and significantly higher precision in speaker diarisation over prior works; 
(ii) an extension of the PopManga evaluation dataset, which now includes annotations for speech-bubble tail boxes, associations of text to corresponding tails, classifications of text as essential or non-essential, and the identity for each character box; and
(iii) a new character bank dataset, which comprises over 11K characters from 76 manga series, featuring 11.5K exemplar character images in total, as well as a list of chapters in which they appear.
The code, trained model, and both datasets can be found at: \url{https://github.com/ragavsachdeva/magi}
\end{abstract}
\vspace{-1em}
\begin{figure}[h]
    \centering
    \vspace{-2em}
    \includegraphics[width=\columnwidth]{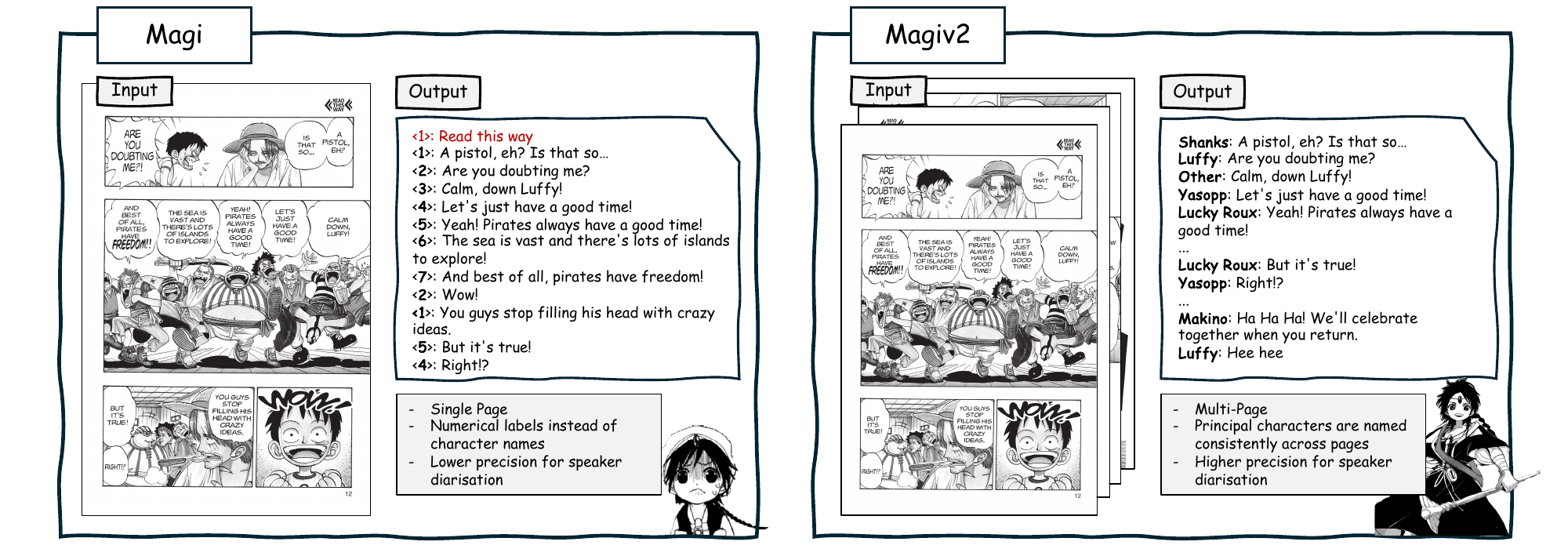}
    \caption{\textbf{(Left)} Magi~\cite{Sachdeva24} generates a page-level transcript, with non-essential texts and without character names.
    \textbf{(Right)} Magiv2 (ours) generates chapter-wide transcripts with principal characters consistently named across pages, higher precision for speaker diarisation and only dialogue-essential texts.}
    \vspace{-4em}
    \label{fig:teaser}
\end{figure}
\section{Introduction}
\noindent Manga, a Japanese form of comic art, is celebrated globally for its rich narratives and distinctive graphical style. It engages millions of readers through its compelling visuals and complex character development. However, this visually dependent medium poses significant accessibility challenges for people with visual impairments (PVI).
Recent advances in computer vision and machine learning present an opportunity to bridge this gap.

Despite the potential, there has been limited research in the field of improving manga accessibility for visually impaired readers. One notable recent work by Sachdeva and Zisserman~\cite{Sachdeva24} addresses the problem of automatically generating transcriptions for manga images. Their contributions include the development of Magi, a model capable of processing a high-resolution manga page to detect characters, texts, and panels, as well as predict character clusters and associate dialogues to their respective speakers. Additionally, they introduced two new datasets, Mangadex-1.5M and PopManga, for training and evaluation purposes.

While the Magi model represents a promising initial step, it remains far from being practically usable due to several critical limitations.
First, a major shortcoming is its failure to incorporate character names within the generated transcripts, instead denoting different characters with numerical labels such as 1, 2, 3, etc.
As the transcripts are generated on a page-by-page basis, this approach inevitably leads to inconsistent character numbering across different pages.
To improve the readability, it is essential to generate chapter-wide transcripts with consistent character names, since numerical labels are non-intuitive and make the transcripts difficult to follow. 
Second, Magi struggles to reliably associate text with the appropriate speaker, often attributing dialogues to the wrong characters. This misattribution disrupts the flow of conversation, leading to a disjointed and confusing narrative. Improving the association of text with the correct speaker is crucial to maintain the coherence of the dialogue and prevent reader confusion.
Third, it is inept at distinguishing between essential and non-essential texts for the dialogue. Non-essential text, such as scene descriptions (e.g., street signs, graffiti, product labels) and sound effects (e.g., ``Thud,'' ``Whoosh''),
should not be attributed to any character and, if improperly included as dialogues in the transcript, can disrupt the narrative flow.

To address these three limitations, we propose Magiv2, a robust and enhanced model capable of generating chapter-wide manga transcripts with consistent character names.
\cref{fig:teaser} shows the comparison of our model with Magi~\cite{Sachdeva24}.
Our approach is built upon several key insights. First, recognising the challenges of character naming in manga transcripts—both providing names and ensuring their consistency—Magiv2 leverages a character bank featuring names and images of principal characters and utilises a training-free, constraint optimisation method to consistently name all characters across the entire chapter, significantly outperforming traditional clustering-based methods.
Second, we note that speech-bubble tails are crucial visual cues, intended by manga artists to indicate who is speaking. By making our model tail-aware, we significantly enhance text-to-speaker association performance.
Third, we observe that distinguishing between essential and non-essential texts can largely be accomplished visually due to differences in font styles and the placement of texts. We leverage this prior and introduce a lightweight text-classification head on top of our visual backbone that can effectively differentiate dialogues from non-dialogue texts.

In summary, we make the following contributions:
(i) We introduce a state-of-the-art model, Magiv2, capable of generating comprehensive manga transcripts across entire chapters, complete with character names and enhanced speaker associations.
(ii) We extend the PopManga evaluation dataset by incorporating annotations for character names, speech bubble tails, text-to-tail associations, and text classification. This dataset is used to evaluate the performance of the new capabilities of Magiv2, such as character recognition.
(iii) We release a new, meticulously curated character bank dataset for 76 manga series, encompassing more than 11K principal characters, with 11.5K exemplar character images in total. Additionally, this dataset includes detailed metadata such as the series names and specific chapters where each character appears. 

With these contributions, over 10,000 published manga chapters (from series in PopManga) can now be transcribed directly using our model, with the potential for more as the character bank dataset grows.
\section{Related Work}
\label{sec:related-work}
\paragraph{Comic understanding.} Using computer vision to analyse and understand comic books has been extensively explored. Several datasets have been contributed to facilitate this research including Manga109~\cite{manga109, manga109coo, manga109dialog}, DCM~\cite{nguyen2018digital}, eBDtheque~\cite{guerin2013ebdtheque}, PopManga~\cite{Sachdeva24} etc. There are several existing works that propose solutions for panel detection~\cite{pang2014robust, ogawa2018object, he2018end, wang2015comic, nguyen2019comic, rigaud2015speech, Sachdeva24}, text/speech balloon detection~\cite{nguyen2019comic, piriyothinkul2019detecting, ogawa2018object, manga109coo, Sachdeva24}, depth-estimation~\cite{bhattacharjee2022estimating}, character detection~\cite{topal2022domain, ogawa2018object, inoue2018cross, jiang2022decoupled, Sachdeva24}, character re-identification, \cite{tsubota2018adaptation, qin2019progressive, zhang2022unsupervised, soykan2023identity, Sachdeva24}, speaker identification, \cite{rigaud2015speech, Sachdeva24}, captioning~\cite{ramaprasad2023comics}, and transcript generation~\cite{Sachdeva24}.

\paragraph{Person identification using a character bank.} Identifying and naming people in images or videos has been a long studied research problem. Often times this either requires complex reasoning, e.g.\ inferring the name of a person based on how other people address them, or prior context and memory, e.g.\ a person may have been introduced previously and this information needs to be remembered. Given the complexity of this task, a common approach is to rely on an external character bank which can be used for matching query character images with a gallery~\cite{Nagrani17b, liu2023detecting, xu2010cast2face, chen2015improving, huang2020movienet, bain2020condensed, hong2023visual}.

\paragraph{Comic Accessibility for PVI users.} Several efforts have been made to understand the challenges faced by PVI when accessing comics~\cite{accesscomics, rayar2017accessible, samarawickrama2023comic} and solutions have been proposed in the form of tactile books~\cite{life_philipp_meyer}, textured images~\cite{textured_images}, audio-books~\cite{star_wars_audiobook} etc. Recent works have also explored the use of computer vision and machine learning to automatically caption simple comic strips~\cite{ramaprasad2023comics} and generate dialogue transcripts of more complex ones~\cite{Sachdeva24}.
\section{Overview of the Chapter-Wide Inference}
\label{sec:method}

\begin{figure*}[!t]
    \centering
    \includegraphics[width=\linewidth]{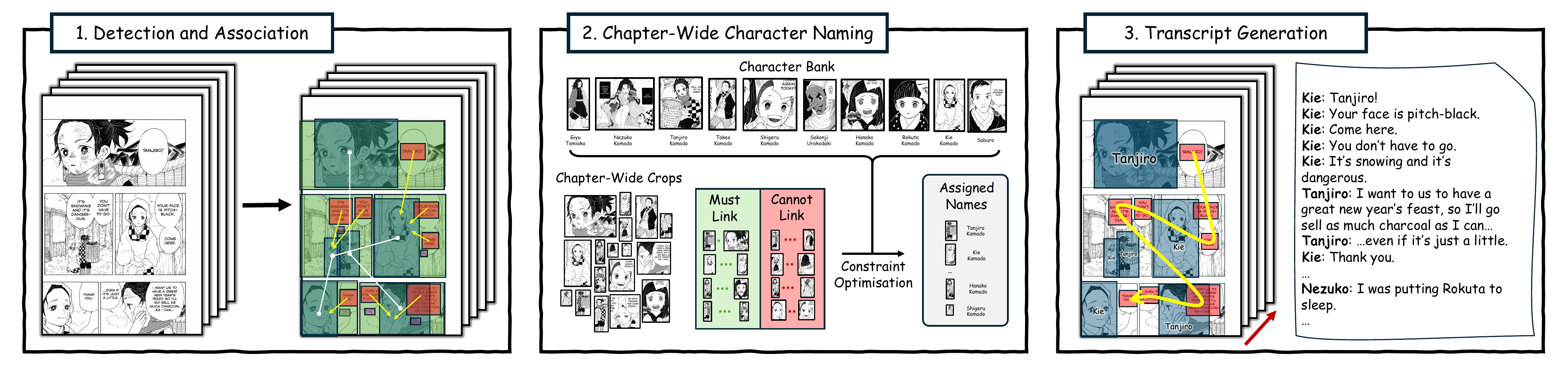}
    \caption{\textbf{Inference pipeline.} Given a manga chapter, along with a character bank: (1) each page is processed independently to detect various elements and their relationships, such as character and text boxes, and their association. Next, (2) using the character bank, names are assigned to detected character crops across all pages using a constraint optimisation approach. Finally, (3) the transcript is generated by performing OCR, ordering all the texts and removing non-essential texts.
    }
    \label{fig:method}
\end{figure*}

\noindent Given a manga chapter, our goal is to generate a transcript of all pages while ensuring narrative consistency.
However, processing all pages simultaneously to generate a dialogue transcript in a single forward pass is computationally prohibitive (a typical manga chapter comprises 15 to 30 pages), necessitating a segmented approach.
To mitigate the computational burden and efficiently generate a chapter-wide transcript with accurate speaker attribution, we employ the following three-step process. The complete inference pipeline is shown in~\cref{fig:method}.

\paragraph{1.\ Detection and Association.}

The initial step involves processing each manga page independently, framed as a graph generation problem.
This step is similar to~\cite{Sachdeva24}, but with modifications to incorporate additional elements.
In our graph, ``nodes'' represent bounding boxes of detected characters, texts, panels, and notably, speech bubble tails. The ``edges'' represent pairwise relationships between character-character, text-character, and text-tail. More details regarding the architecture and training strategy are described in~\cref{sec:detection_and_association}.

\paragraph{2.\ Chapter-Wide Character Naming.}
Given the crops of detected characters from all pages of a manga chapter, along with a character bank comprising images and names of principal characters, the goal is to assign each character crop to the correct principal character in the character bank, if one exists, otherwise assign it to the ``other'' class. This step simultaneously allows naming of speakers in the final transcripts, and consistent character identification across pages, if the assignments are correct. We formulate this as a constraint optimisation approach and provide more details in~\cref{sec:character_naming}.

\paragraph{3.\ Transcript Generation.}
Finally, the gathered information is compiled to generate the chapter-wide transcript. This is a four step process: First, the detected text boxes are filtered such that the texts that are classified as non-essential are removed; Second, the remaining text boxes are organised in their reading order (by first sorting the pages, then sorting panels on each page~\cite{Sachdeva24} and finally sorting texts within each panel~\cite{hinami2021towards}); Third, Optical Character Recognition (OCR) is used to extract texts from the manga pages; Finally, the transcript is generated by utilising the text-character associations predicted in~\cref{sec:detection_and_association} and character names predicted in~\cref{sec:character_naming}. We provide more details in supp. mat.

\section{Detection and Association}
\label{sec:detection_and_association}
\begin{figure}[!t]
    \centering
    \includegraphics[width=\textwidth]{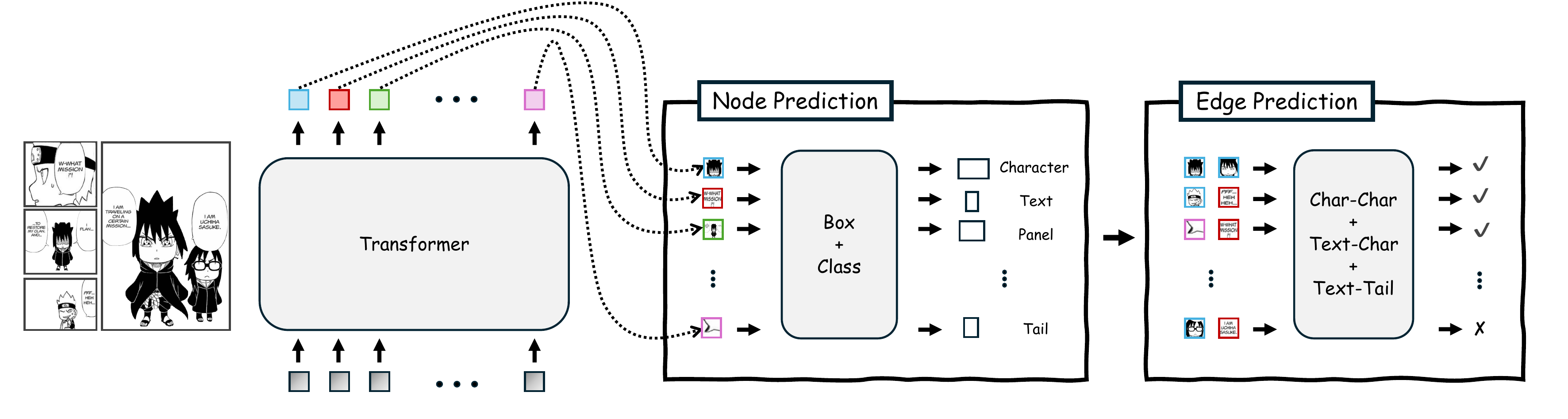}
    \caption{\textbf{Simplified Detection and Association Architecture}. The input to the model is an RGB image of a manga page. The transformer decoder outputs several feature vectors which are used to predict bounding boxes for characters, texts, panels and tails (``nodes'').
    These features are further processed in pairs to predict character-character, text-character and text-tail associations (``edges'').
    }
    \label{fig:arch}
\end{figure}

Given a manga page, the objective of this section
is to detect the various components that constitute a manga page---particularly the panels, characters, text blocks and tails (i.e.,\ localise where they are on the page), and also to associate them:
character-character
association (i.e.,\ character clustering), text-character association (i.e.\ speaker diarisation), and text-tail association. We cast this as generating a graph, as
described in step 1 of the inference process of~\cref{sec:method}. The model architecture for this detection and association is illustrated in~\cref{fig:arch}.

Briefly, the model ingests a high resolution manga page as input, which is first processed by a CNN backbone, followed by a transformer encoder-decoder resulting in $N$ object feature vectors. These feature vectors are processed by the detection head to regress a bounding box and classify it into character, text, panel, tail or background.
This completes the \textit{nodes} part of the graph. To generate text-character \textit{edges}, the features corresponding to detected text boxes and character boxes are processed in pairs by a speaker association head to make a binary prediction (whether the edge exists or not). Similarly, a tail association head processes pairs of text and tail feature vectors resulting in text-tail \textit{edges}. Character-character \textit{edges} are obtained by processing pairs of detected character feature vectors along with their respective crop embedding feature vectors (obtained by a separate crop embedding module). 
Finally, the feature vectors corresponding to detected texts are processed by a linear layer to classify them into essential vs non-essential.
Further details regarding the model architecture and implementation are provided in \cref{sec:implementation}.

\subsection{Semi-Supervised Training} 

A significant challenge in training the graph generation model is the quality and completeness of the training data. 
We utilise two datasets: (i) Mangadex-1.5M, which is unlabelled, and (ii) PopManga (Dev), which is partially labelled\footnote{\label{partially_annotated} All pages contain character boxes, text boxes, and character clusters, but only a subset of pages include text-to-character associations. Furthermore, none of the pages have annotations for speech bubble tails or labels for text classification.}.

To address the challenge of partially annotated training data, we approach the training of our graph generation model through semi-supervised learning. We begin by curating a small subset of the PopManga (Dev) set, and endow it with both tail-related annotations and text classification labels. Additionally, we extract \textit{partial} labels for the Mangadex-1.5M dataset using Magi~\cite{Sachdeva24}, which of course 
lack tail-related and text classification annotations. Given this combination of data---small subset with complete annotations and larger subsets with partial pseudo-annotations---we adopt the following training strategy.

Initially, we warm up our model by training it on the large-scale partially-labeled pseudo-annotations. We then fine-tune this model on the smaller, fully annotated dataset. Subsequently, this model is used to mine \textit{complete}, but possibly noisy, annotations for the large-scale data. We then re-train our model \textit{from scratch} on this newly annotated large-scale data, which remains pseudo-annotated but now more comprehensively annotated, followed by additional fine-tuning on the small, completely annotated data. This cycle is repeated multiple times to refine our model. Detailed training recipe is provided in the supp.\ mat.

This training approach ensures robust detection and association capabilities despite the incomplete initial annotations. This style of training paradigm is common in noisy label learning literature, where training the model on noisy but large scale data (in our case pseudo-annotations), followed by fine-tuning on clean data, and then re-mining the pseudo-annotations, results in improved training data, which in turn can be used to train a better model~\cite{arazo2019unsupervised, li2020dividemix, sachdeva2021evidentialmix, sachdeva2023scanmix, cordeiro2023longremix}. A crucial aspect of this methodology is the re-training of the model \textit{from scratch} after each round of mining pseudo-annotations, which is essential to avoid confirmation bias and prevent the model from overfitting on its own predictions.

\subsection{Implementation}
\label{sec:implementation}
The graph generation model architecture 
consists of a ResNet50~\cite{he2016deep} backbone, followed by a encoder-decoder transformer with 6 layers each, hidden dimension of $256$, $8$ attention heads and conditional cross-attention~\cite{meng2021conditional}. The crop-embedding module is an encoder-only transformer with 12 layers, hidden dimension of 768, and 12 attention heads. The text-character, text-tail and character-character edge prediction heads are all 3-layered MLPs, and the text-classification head is a simple linear layer. Our training objective for box prediction is the same as in~\cite{meng2021conditional}. We further apply Binary Cross Entropy loss to the outputs of our edge-prediction as well as text-classification heads. Additionally, we apply Supervised Contrastive Loss~\cite{khosla2020supervised} to the per-page embeddings from the crop-embedding module.
We trained our model, on $2\times$A40 GPUs using AdamW~\cite{loshchilov2017decoupled} optimiser with both learning rate and weight decay of 0.0001, and batch size of 16.

\section{Chapter-Wide Character Naming}
\label{sec:character_naming}
The objective in this section to {\em assign} each detected character crop in the chapter to one of the characters in the character bank (introduced in~\cref{sec:datasets}), unless they are ``other''. This is the second, chapter-wide, step of the inference process. 

The question is, how to optimise this assignment objective? Naively, this can be accomplished greedily by computing the similarity of each crop with each character in the character bank and taking the \texttt{argmax}. However, we can do better, by leveraging additional \textit{constraints} (must-link and cannot-link) from per-page associations. Specifically, the graph computed in~\cref{sec:detection_and_association}, provides us with character-character edges which can be transformed into must-link constraints, i.e.\ these crops are of the same characters and must be assigned the same identity, and cannot-link constraints, i.e.\ these crops are of different characters and must be assigned to different identities. These per-page must-link and cannot-link constraints provide a stronger signal than simple crop-based similarity as they factor in surrounding visual cues, e.g.\ two characters in the same panel are likely to be different characters, regardless of the visual similarity. This assignment problem can be formulated as a Mixed Integer Linear Programming problem, for which there are several existing solvers e.g.\ COIN-OR Branch and Cut Solver (CBC)~\cite{cbc,coin-or}.\newline

\noindent\textit{Problem Definition.} Formally, suppose there are $n$ character crops in a particular chapter and $k$ characters in the character bank. Additionally, suppose that we have a set of must-link constraints $M$, representing pairs of crops that must be assigned to the same character, and a set of cannot-link constraints $C$, representing pairs of crops that must not be assigned to the same character. We further define a $(k+1)$th character, which is a dummy character to capture ``other'' when the crop is of a character that is not in the character bank.

\noindent\textit{Variable.} Let $x_{ij}$ be a binary variable that equals 1 if character crop $i$ is assigned to character $j$ in the character bank, and 0 otherwise.

\noindent\textit{Objective Function.} The objective is to compute the optimal assignment of crops to characters in the character bank, i.e.\ computing $x_{ij}$, which is achieved by

\begin{equation}
\min_x \sum_{i=1}^{n} \sum_{j=1}^{k+1} d_{ij} x_{ij}, \quad \text{where} \quad d_{ij} = \begin{cases}
\eta  & \text{if } j = k+1,\\
\|e_i - e_j\| & \text{otherwise}
\end{cases}
\end{equation}
 
 and \( e_i, e_j \) are embeddings for crop \( i \) and  character \( j \), respectively, and $\eta$ is a fixed outlier-threshold hyperparameter (in practice, $\eta =$ 0.75).

\noindent\textit{Constraints.} The objective function above is minimised subject to the following constraints:
\begin{equation}
\sum_{j=1}^{k+1} x_{ij} = 1, \quad \forall i \in \{1, \ldots, n\}
\label{eq:one_only}
\end{equation}
\begin{equation}
x_{u,j} - x_{v,j} = 0, \quad \forall (u, v) \in M, \quad \forall j \in \{1, \ldots, k+1\}
\label{eq:must_link}
\end{equation}
\begin{equation}
x_{u,j} + x_{v,j} \leq 1, \quad \forall (u, v) \in C, \quad \forall j \in \{1, \ldots, k\}
\label{eq:cannot_link}
\end{equation}

where~\cref{eq:one_only} ensures that each crop is assigned to exactly one character,~\cref{eq:must_link} enforces the must-link constraints, and~\cref{eq:cannot_link} enforces the cannot-link constraints. 

Note that in~\cref{eq:cannot_link}, $j\neq k+1$. This is because there may be two different characters (hence \textit{must not be} linked) that are not in the character bank (hence \textit{must be} linked to ``other''). In other words, a cannot link constraint between crops $u, v$ is applied such that these crops must not be assigned to the same character, unless they are assigned to ``other'' i.e.\ $(k+1)$th character.

\noindent In~\cref{sec:results}, we compare the proposed constraint optimisation approach with traditional clustering based approaches and demonstrate that our method significantly outperforms the baselines.
\begin{figure*}[!t]
    \centering
    \includegraphics[width=\linewidth]{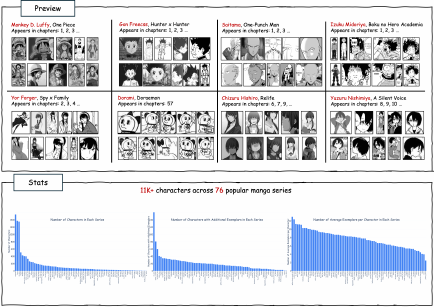}
    \caption{\textbf{\cbd---the proposed character bank dataset.} \textbf{(Top)} We show exemplars for eight different characters from \cbd, along with their name (in red), followed by the name of the series and the list of chapters they appear in. \textbf{(Bottom)} We display histograms showing the number of  characters (left), the number of frequently-occurring characters with additional exemplar images (middle), and the average number of exemplars per frequently-occurring character (right) in each series for \cbd. 
    }
    \label{fig:popcharacters}
\end{figure*}

\section{Datasets: \cbd and \pmx}
\label{sec:datasets}

The recently introduced PopManga~\cite{Sachdeva24} dataset provides annotations for character boxes, text boxes, per-page character clusters and speaker associations. It  is divided into three splits: Dev, Test-S and Test-U, with around 2000 images of manga pages in Test, and S \& U meaning that other chapters from the series are Seen or Unseen during training.

\begin{figure}[!t]
    \centering
    \includegraphics[width=0.9\textwidth]{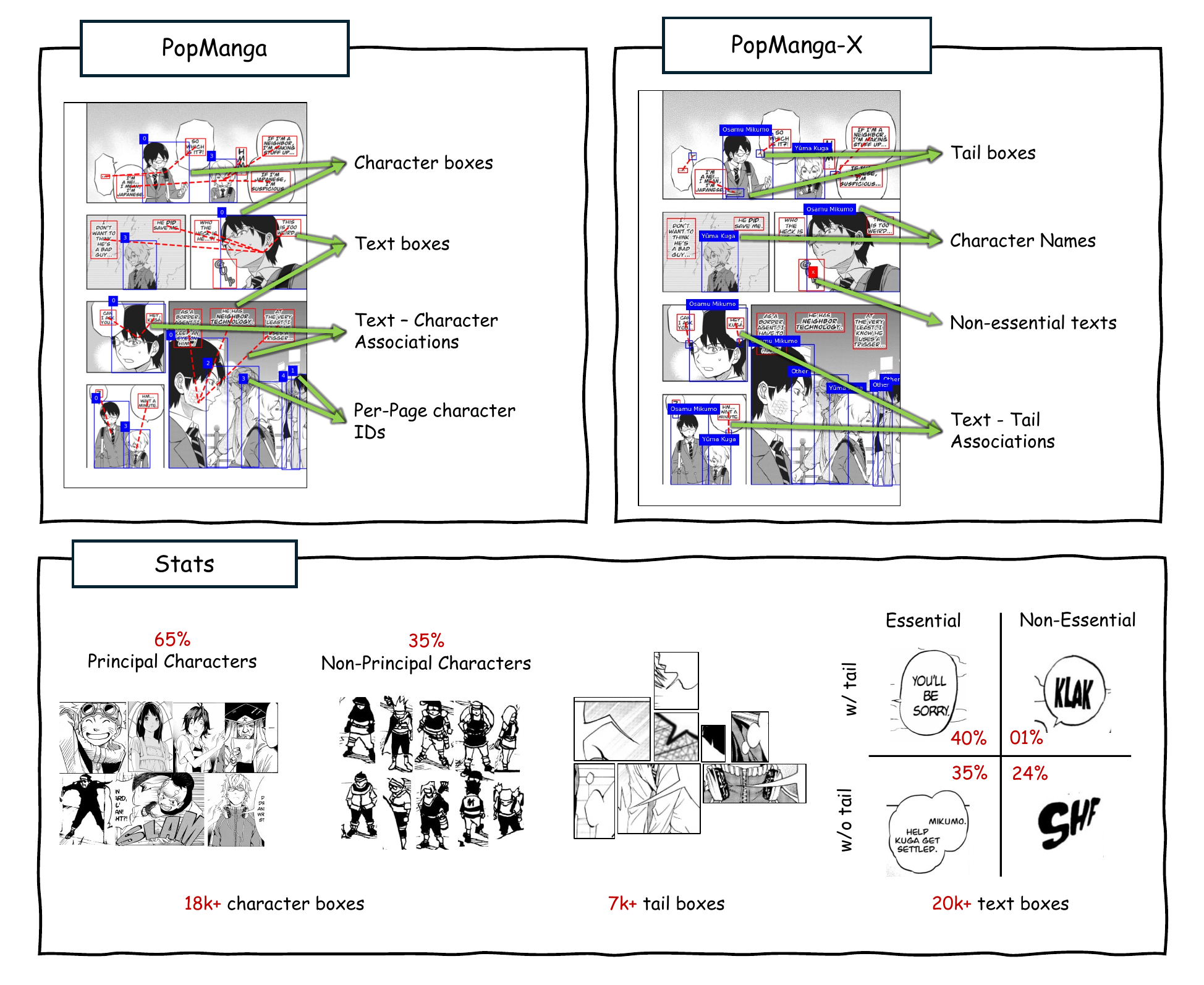}
    \caption{\textbf{\pmx}. \textbf{(Top-Left)} Ground-truth annotations in PopManga. \textbf{(Top-Right)} Ground-truth annotations \textit{added} to test splits of PopManga, now referred to as PopManga-X. \textbf{(Bottom)} Statistics on various elements of PopManga-X test splits.
    }
    \label{fig:popmanga-test-x}
\end{figure}

In this section we detail two data related contributions: (a) We compile a character bank of principal manga characters in PopManga.  Please see~\cref{fig:popcharacters} for some examples and dataset statistics; (b) We extend the annotations of
the PopManga test set to facilitate the evaluation of the new Magiv2 model capabilities, such as character labelling.
Please see~\cref{fig:popmanga-test-x} for an overview of the extended test dataset along with statistics on various types of available annotations.\newline

\mypara{\cbd.}
We introduce a new character bank dataset, called \cbd, comprising principal characters\footnote{We define principal characters as those who play crucial roles in the main story of a series. Please refer to the supplementary material for more details.} in PopManga. For each principal character in \cbd, we provide (i) the character's name, (ii) a set of web-scraped thumbnail images of the character, and (iii) the series it belongs and a list of manga chapters the character appears in. Additionally, for a subset of the characters in PopCharacters, which appear far more frequently than others, we also provide a set of exemplar images queried from within the manga chapters and verified by human-in-the-loop.
The purpose of this dataset is two-fold:
(i) it enables Magiv2 to transcribe hundreds of thousands of manga pages (that are commercially available), with names for principal characters; and
(ii) it provides a valuable resource for training models on tasks such as character recognition and character clustering. Further details on the dataset curation process is provided in the supp.\ mat.

\mypara{\pmx.}
In the PopManga  test splits (i.e., Test-S and Test-U), we provide new annotations for speech-bubble tail bounding boxes, associations of text boxes to tail boxes, and text categories (essential vs non-essential), providing the test-bed for tail-related predictions and text classification (see \cref{fig:popmanga-test-x}).
Moreover, we label each character box in the test splits with the name of the character (consistent with PopCharacters), thereby offering global character clusters across the series and permitting the evaluation of chapter-wide character cluster predictions.
To differentiate this extended dataset with more types of annotations from the original, we call this \pmx. Further details are provided in supp. mat.

\section{Results}
\label{sec:results}

\begin{figure}
    \centering
    \includegraphics[width=\columnwidth]{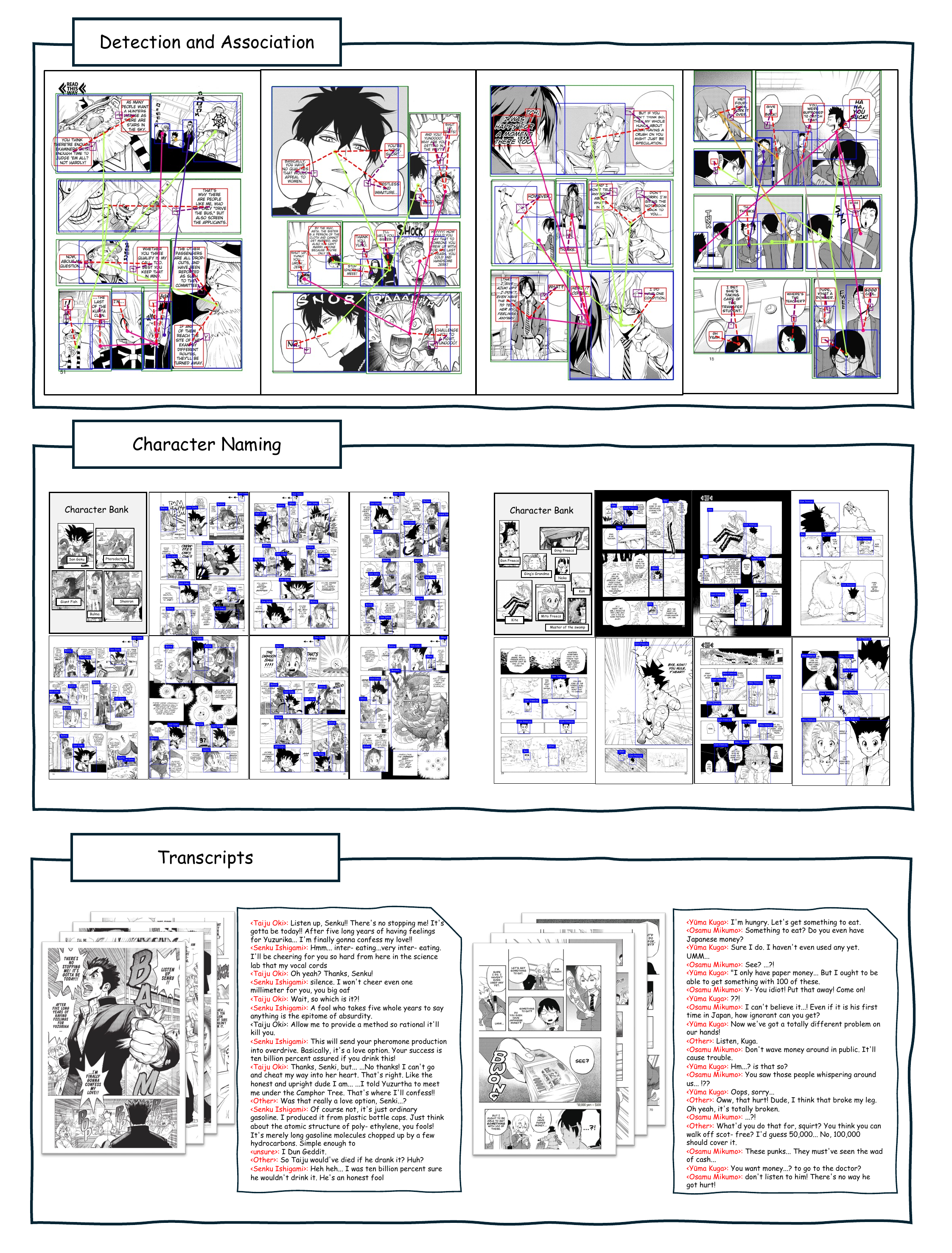}
    \caption{\textbf{Qualitative Predictions.} \textbf{(Top)} We show the predictions from our graph generation model---panels (in green), character (in blue), texts classified as essential (in red), and speech bubble tails (in purple). The text-character edges (dashed red lines), text-tail edges (dashed purple lines) and character-character edges (unique colour for each connected component) are also shown. \textbf{(Middle)} We show the prediction for character names across multiple pages of two different manga series, demonstrating character naming consistency. \textbf{(Bottom)} We show the final, generated, multi-page transcripts using our method.
    }
    \label{fig:qualitative_predictions}
\end{figure}

To achieve high-quality chapter-wide manga transcriptions, we identify three core tasks in~\cref{sec:method}: (i) per-page detection and association, (ii) chapter-wide character naming, and (iii) transcript generation.
In the following, we report our model's performance on the first two tasks and note that their quality directly determines the quality of the third. In other words, to evaluate the quality of the generated transcript, it is enough to evaluate the first two tasks only, as they reflect the correctness of the transcript. 
Qualitative results are shown in~\cref{fig:qualitative_predictions}.

\subsection{Detection and Association/Graph Generation}
Given a single manga page as input, here we evaluate the performance of the graph generation model in terms of (i) predicting panels, texts, characters, and tails, (ii) predicting text-character edges, text-tail edges, character-character edges, and (iii) classifying predicted texts into essential vs non-essential.\newline

\mypara{Nodes:} To measure the quality of predicted ``nodes'' i.e.\ predicted bounding boxes, we use the standard object detection evaluation measures. In~\cref{tab:detection}, we report the results for the average precision metric~\cite{lin2014microsoft} for detecting character boxes, text boxes, tail boxes and panel boxes. We compare our results against, Magi~\cite{Sachdeva24}, DASS~\cite{topal2022domain} and zero-shot results from GroundingDino~\cite{grounding_dino}, on \pmx and Manga109~\cite{manga109}.\newline

\mypara{Edges:} Our model is trained to predict three different kinds of edges: (i) character-character, (ii) text-character and, (iii) text-tail.
To evaluate the character-character edges, we treat it as a per-page clustering problem and report the same metrics as~\cite{Sachdeva24}, namely AMI, NMI, P@1, R-P, MRR and MAP@R, on \pmx and Manga109~\cite{manga109}, in~\cref{tab:char_clus}.
For text-character and text-tail edges, we treat it as a binary classification problem (whether the predicted edge is correct or not) and report the average precision metric on \pmx in~\cref{tab:diarisation}.
We compare our results against~\cite{Sachdeva24} and provide more details on the edge evaluation procedure in the supp.\ mat.\newline

\setlength{\tabcolsep}{4pt}
\renewcommand{\arraystretch}{1.0}

\begin{table}[H]
\caption{\textbf{Detection Results.} We report the average precision results, which have an upper bound of 1.0.
}
\centering
\resizebox{\columnwidth}{!}{%
\begin{tabular}{lccc|ccc|cc}
                                                          & \multicolumn{3}{c|}{\pmx (Test-S)}                                                        & \multicolumn{3}{c|}{\pmx (Test-U)}                                                        & \multicolumn{2}{c}{Manga109}                           \\ \hline
\multicolumn{1}{l|}{method}                               & \multicolumn{1}{c|}{Char}            & \multicolumn{1}{c|}{Text}            & Tail            & \multicolumn{1}{c|}{Char}            & \multicolumn{1}{c|}{Text}            & Tail            & \multicolumn{1}{c|}{Body}            & Panel           \\ \hline
\multicolumn{1}{l|}{DASS~\cite{topal2022domain}}          & \multicolumn{1}{c|}{0.8410}          & \multicolumn{1}{c|}{-}               & -               & \multicolumn{1}{c|}{0.8580}          & \multicolumn{1}{c|}{-}               & -               & \multicolumn{1}{c|}{\textbf{0.9251}} & -               \\
\multicolumn{1}{l|}{Grounding-DINO~\cite{grounding_dino}} & \multicolumn{1}{c|}{0.7250}          & \multicolumn{1}{c|}{0.7922}          & -               & \multicolumn{1}{c|}{0.7420}          & \multicolumn{1}{c|}{0.8301}          & -               & \multicolumn{1}{c|}{0.7985}          & 0.5131          \\
\multicolumn{1}{l|}{Magi~\cite{Sachdeva24}}               & \multicolumn{1}{c|}{0.8485}          & \multicolumn{1}{c|}{0.9227}          & -               & \multicolumn{1}{c|}{0.8615}          & \multicolumn{1}{c|}{0.9208}          & -               & \multicolumn{1}{c|}{0.9015}          & 0.9357          \\
\multicolumn{1}{l|}{Magiv2 (Ours)}                      & \multicolumn{1}{c|}{\textbf{0.8544}} & \multicolumn{1}{c|}{\textbf{0.9372}} & \textbf{0.8766} & \multicolumn{1}{c|}{\textbf{0.8720}} & \multicolumn{1}{c|}{\textbf{0.9353}} & \textbf{0.8737} & \multicolumn{1}{c|}{0.9046}          & \textbf{0.9405}
\end{tabular}
}
\label{tab:detection}
\end{table}

\mypara{Text Classification:} Finally, we evaluate the performance of our model on categorising the detected texts into essential vs non-essential. In this work, we classify a text as essential, if it is a spoken dialogue, interjection or an internal thought by a character, or context added by the narrator. Everything else is non-essential e.g.\ sound-effects, editorial footnotes, scene-texts etc. We report the average precision results on \pmx in~\cref{tab:diarisation}. We use the confidence scores from Magi~\cite{Sachdeva24} as a baseline, while acknowledging that it was not explicitly trained for this task.\newline

\mypara{Discussion.} When compared with prior works, our model achieves (i) better per-page character clustering results, particularly in the crop-only setting, which we attribute to the semi-supervised learning training scheme, where mining better pseudo labels in turn improves the model's performance; (ii) significant improvement in speaker diarisation (i.e.\ text-character matching) results, which is largely attributed to the introduction of speech-bubble tails; (iii) comparable bounding box detection results. Furthermore, our work unlocks new functionality, not supported by prior works, including (i) detecting tail boxes, text-tail associations, and (ii) text classification into essential vs non-essential, which can be used to improve the quality of the generated transcripts.

\setlength{\tabcolsep}{4pt}

\renewcommand{\arraystretch}{1.0}
\begin{table}[!t]
\footnotesize
\caption{\textbf{Per-Page Character Clustering Results.} We report results using several metrics. They all have an upper bound of 1.0. 
}
\centering

\begin{tabular}{lcccccc}
\hline
method                                                  & AMI                                  & NMI                                  & MRR                                  & MAP@R                                & P@1                                  & R-P             \\ \hline
                                                        & \multicolumn{6}{c}{\pmx (Test-S)}                                                                                                                                                                                  \\ \hline
\multicolumn{1}{l|}{Magi (crop only)~\cite{Sachdeva24}} & \multicolumn{1}{c|}{0.4892}          & \multicolumn{1}{c|}{0.7178}          & \multicolumn{1}{c|}{0.9008}          & \multicolumn{1}{c|}{0.7840}          & \multicolumn{1}{c|}{0.8423}          & 0.8008          \\
\multicolumn{1}{l|}{Magiv2 (crop only) (Ours)}          & \multicolumn{1}{c|}{\textbf{0.5826}} & \multicolumn{1}{c|}{\textbf{0.8120}} & \multicolumn{1}{c|}{\textbf{0.9275}} & \multicolumn{1}{c|}{\textbf{0.8401}} & \multicolumn{1}{c|}{\textbf{0.8831}} & \textbf{0.8526} \\ \hline
\multicolumn{1}{l|}{Magi~\cite{Sachdeva24}}             & \multicolumn{1}{c|}{0.6574}          & \multicolumn{1}{c|}{0.8501}          & \multicolumn{1}{c|}{0.9312}          & \multicolumn{1}{c|}{0.8439}          & \multicolumn{1}{c|}{0.8884}          & 0.8555          \\
\multicolumn{1}{l|}{Magiv2 (Ours)}                      & \multicolumn{1}{c|}{\textbf{0.6745}} & \multicolumn{1}{c|}{\textbf{0.8610}} & \multicolumn{1}{c|}{\textbf{0.9431}} & \multicolumn{1}{c|}{\textbf{0.8669}} & \multicolumn{1}{c|}{\textbf{0.9066}} & \textbf{0.8770} \\ \hline
                                                        & \multicolumn{6}{c}{\pmx (Test-U)}                                                                                                                                                                                  \\ \hline
\multicolumn{1}{l|}{Magi (crop only)~\cite{Sachdeva24}} & \multicolumn{1}{c|}{0.4862}          & \multicolumn{1}{c|}{0.7326}          & \multicolumn{1}{c|}{0.9061}          & \multicolumn{1}{c|}{0.7926}          & \multicolumn{1}{c|}{0.8477}          & 0.8076          \\
\multicolumn{1}{l|}{Magiv2 (crop only) (Ours)}          & \multicolumn{1}{c|}{\textbf{0.5711}} & \multicolumn{1}{c|}{\textbf{0.8108}} & \multicolumn{1}{c|}{\textbf{0.9321}} & \multicolumn{1}{c|}{\textbf{0.8491}} & \multicolumn{1}{c|}{\textbf{0.8898}} & \textbf{0.8598} \\ \hline
\multicolumn{1}{l|}{Magi~\cite{Sachdeva24}}             & \multicolumn{1}{c|}{0.6527}          & \multicolumn{1}{c|}{0.8503}          & \multicolumn{1}{c|}{0.9347}          & \multicolumn{1}{c|}{0.8557}          & \multicolumn{1}{c|}{0.8936}          & 0.8656          \\
\multicolumn{1}{l|}{Magiv2 (Ours)}                      & \multicolumn{1}{c|}{\textbf{0.6650}} & \multicolumn{1}{c|}{\textbf{0.8579}} & \multicolumn{1}{c|}{\textbf{0.9508}} & \multicolumn{1}{c|}{\textbf{0.8818}} & \multicolumn{1}{c|}{\textbf{0.9202}} & \textbf{0.8898} \\ \hline
                                                        & \multicolumn{6}{c}{Manga109 (Body)}                                                                                                                                                                           \\ \hline
\multicolumn{1}{l|}{Magi (crop only)~\cite{Sachdeva24}} & \multicolumn{1}{c|}{0.5690}          & \multicolumn{1}{c|}{0.7694}          & \multicolumn{1}{c|}{0.9237}          & \multicolumn{1}{c|}{0.8259}          & \multicolumn{1}{c|}{0.8721}          & 0.8389          \\
\multicolumn{1}{l|}{Magiv2 (crop only) (Ours)}          & \multicolumn{1}{c|}{\textbf{0.6204}} & \multicolumn{1}{c|}{\textbf{0.8152}} & \multicolumn{1}{c|}{\textbf{0.9400}} & \multicolumn{1}{c|}{\textbf{0.8646}} & \multicolumn{1}{c|}{\textbf{0.9002}} & \textbf{0.8737} \\ \hline
\multicolumn{1}{l|}{Magi~\cite{Sachdeva24}}             & \multicolumn{1}{c|}{0.6345}          & \multicolumn{1}{c|}{0.8202}          & \multicolumn{1}{c|}{0.9383}          & \multicolumn{1}{c|}{0.8567}          & \multicolumn{1}{c|}{0.8966}          & 0.8667          \\
\multicolumn{1}{l|}{Magiv2 (Ours)}                      & \multicolumn{1}{c|}{\textbf{0.6456}} & \multicolumn{1}{c|}{\textbf{0.8336}} & \multicolumn{1}{c|}{\textbf{0.9514}} & \multicolumn{1}{c|}{\textbf{0.8812}} & \multicolumn{1}{c|}{\textbf{0.9179}} & \textbf{0.8895} \\ \hline
\end{tabular}
\label{tab:char_clus}
\end{table}

\normalsize
\setlength{\tabcolsep}{4pt}
\renewcommand{\arraystretch}{1.0}

\begin{table}[H]
\caption{\textbf{Text-Related Results.} We report the average precision results, which have an upper bound of 1.0.}
\centering
\resizebox{\columnwidth}{!}{%
\begin{tabular}{l|cc|cc|cc}
                       & \multicolumn{2}{c|}{Text - Character Association}      & \multicolumn{2}{c|}{Text - Tail Association}          & \multicolumn{2}{c}{Text Classification}                \\ \cline{2-7} 
method                 & \multicolumn{1}{c|}{\pmx (Test-S)}   & \pmx (Test-U)   & \multicolumn{1}{c|}{\pmx (Test-S)}   & \pmx (Test-U)  & \multicolumn{1}{c|}{\pmx (Test-S)}   & \pmx (Test-U)   \\ \hline
Magi~\cite{Sachdeva24} & \multicolumn{1}{c|}{0.5248}          & 0.5632          & \multicolumn{1}{c|}{-}               & -              & \multicolumn{1}{c|}{0.9617}          & 0.9692          \\ \hline
Magiv2 (Ours)          & \multicolumn{1}{c|}{\textbf{0.7499}} & \textbf{0.7512} & \multicolumn{1}{c|}{\textbf{0.9838}} & \textbf{0.9830} & \multicolumn{1}{c|}{\textbf{0.9897}} & \textbf{0.9914}
\end{tabular}
}
\label{tab:diarisation}
\end{table}

\subsection{Chapter-Wide Character Naming/Character Identification}
Here we evaluate the efficacy of our method in forming chapter-wide character clusters and evaluate whether the same characters across pages are assigned the same name. For evaluation, we utilise test splits of \pmx where the input this time is an entire manga chapter. There are 50 chapters in the two test sets in total. For each chapter, we curate a chapter-specific character bank from the PopCharacters dataset, comprising principal characters that appear in this chapter. This chapter-specific character bank consists of names of principal characters along with exactly 1 exemplar image per character. Furthermore, all non-principal characters in the chapter, i.e.\ characters for which we do not have a name and exemplar image, are grouped into a single ``other'' category. Given a manga chapter and chapter-specific character bank, we report the accuracy of character naming, in~\cref{tab:character_naming}.\newline

\noindent\textbf{Naive Baseline.} A straightforward solution to the chapter-wide character naming problem is to formulate it as a clustering problem. Assuming that the number of characters in the character bank, $k$, is equal to the number of ground truth clusters in the chapter (which may not be true in practice), a simple approach is to compute embeddings for each character crop in the chapter and then cluster them into $k$ clusters.

We take two approaches to implement clustering based baselines: (i) simple K-means clustering~\cite{lloyd1982kmeans}, with $k+1$ clusters (an extra cluster for ``other'' characters, not in the character bank); and (ii) first filter out all ``other'' characters using outlier/anomaly detection~\cite{liu2012isolation} and then perform simple K-means clustering with $k$ clusters. Once the clusters have been computed they are assigned to character names by using Hungarian matching~\cite{Kuhn1955Hungarian} between the embeddings for cluster centres and exemplar images in the character bank.\newline

\mypara{Discussion.} When compared with traditional clustering based solutions, we show that our method performs significantly better. In particular, there are two critical shortcomings with the clustering-based baselines: i) similar looking, but distinct, characters are very likely to be grouped into a single cluster. This further impacts the cluster assignment of other crops, given that the number of clusters is fixed; (ii) such methods do not leverage spatial cues to assist in clustering, e.g.\ two characters in the same panel are likely to be different characters, regardless of their visual similarity. Our proposed method is more robust to such shortcomings as evident by the superior performance. We also observe that using ground truth must-link and cannot-link constraints for each page, significantly improves the quality of the results.
This finding has a very \textit{\textbf{significant implication}}---in the future, it is sufficient to improve the \textit{per-page} model in order to improve the \textit{chapter-wide} results. This is of great value because training a per-page model is much more tractable.
In the supplementary, we show how the choice of reference exemplar image in the character bank can further impact the performance.
A key \textit{\textbf{limitation}} of this approach, however, is that it groups all non-principal characters into a single ``other'' category. It is not designed to disambiguate `unnamed person 1' from `unnamed person 2'. We leave this as future work.

\setlength{\tabcolsep}{4pt}
\renewcommand{\arraystretch}{1.0}

\begin{table}[H]
\caption{\textbf{Character Naming Results.} We report the accuracy results, which have an upper bound of 1.0.}
\centering
\resizebox{\columnwidth}{!}{%
\begin{tabular}{l|c|c|c|c|}
\rowcolor[HTML]{EFEFEF} embedding model & method                                                           & notes                          & \multicolumn{1}{l|}{\cellcolor[HTML]{EFEFEF}\pmx (Test-S)} & \multicolumn{1}{l|}{\cellcolor[HTML]{EFEFEF}\pmx (Test-U)} \\ \hline
Magi                                    & K-means~\cite{lloyd1982kmeans}                                   & nclusters$=k+1$                & 0.3351                                                     & 0.3820                                                     \\ \hline
Magiv2                                  & K-means~\cite{lloyd1982kmeans}                                   & nclusters$=k+1$                & 0.3801                                                     & 0.4223                                                     \\ \hline
Magi                                    & iForest~\cite{liu2012isolation} + K-means~\cite{lloyd1982kmeans} & nclusters$=k$                  & 0.4549                                                     & 0.4646                                                     \\ \hline
Magiv2                                  & iForest~\cite{liu2012isolation} + K-means~\cite{lloyd1982kmeans} & nclusters$=k$                  & 0.5101                                                     & 0.4942                                                     \\ \hline
Magi                                    & Constraint Optimisation (Ours)                                   & Predicted per-page constraints & 0.6637                                                     & 0.7058                                                     \\ \hline
Magiv2                                  & Constraint Optimisation (Ours)                                   & Predicted per-page constraints & \textbf{0.7273}                                            & \textbf{0.7530}                                            \\ \hline
\rowcolor[HTML]{EFEFEF} Magi            & Constraint Optimisation (Ours)                                   & GT per-page constraints        & \cellcolor[HTML]{EFEFEF}0.7445                             & \cellcolor[HTML]{EFEFEF}0.7975                             \\ \hline
\rowcolor[HTML]{EFEFEF} Magiv2          & Constraint Optimisation (Ours)                                   & GT per-page constraints        & \cellcolor[HTML]{EFEFEF}0.7987                             & \cellcolor[HTML]{EFEFEF}0.8526                            
\end{tabular}
}
\label{tab:character_naming}
\end{table}

\section{Conclusion}
In this work, we present a solution for generating chapter-wide manga transcriptions with consistent character names and clearer narrative. We contribute a new SOTA model, a training-free constraint optimisation approach to chapter-wide character naming, and new datasets to facilitate further research and comparisons. With these contributions it is now possible to transcribe over 10,000 manga chapters that are currently available commercially, complete with character names, allowing for a much richer reading experience for the visually impaired audience. 

\newpage
\noindent \textbf{Acknowledgements:} This research is supported by EPSRC Programme Grant VisualAI EP/T028572/1 and a Royal Society Research Professorship RP\textbackslash R1\textbackslash191132.
This work was partially supported using resources provided by the Cambridge Service for Data Driven Discovery (CSD3) operated by the University of Cambridge Research Computing Service (www.csd3.cam.ac.uk), provided by Dell EMC and Intel using Tier-2 funding from the Engineering and Physical Sciences Research Council (capital grant EP/T022159/1), and DiRAC funding from the Science and Technology Facilities Council (www.dirac.ac.uk).
Gyungin Shin would like to thank Zheng Fang for the enormous support.

%
%
\bibliographystyle{splncs04}
\bibliography{main,vgg_local}

\title{Tails Tell Tales: Chapter-Wide Manga Transcriptions with Character Names\\
--- Supplementary Material}

\titlerunning{Tails Tell Tails}

\author{Ragav Sachdeva \and
Gyungin Shin \and
Andrew Zisserman}

\authorrunning{Sachdeva et al.}

\institute{Visual Geometry Group, Dept. of Engineering Science, University of Oxford}

\maketitle

\section{More on datasets}
In this section we provide more details regarding the two dataset contributions---PopCharacters and PopManga-X.

\subsection{PopCharacters}
PopCharacters is a character bank dataset comprising principal characters from PopManga dataset, and containing information such as the characters' names, the manga series each character belongs to, a list of chapters that each character appears in, and a  set of exemplar images for each character. In the following we provide details on the data curation process.

\subsubsection{Web-scraping.}

\begin{figure*}[h]
    \centering
    \includegraphics[width=0.8\linewidth]{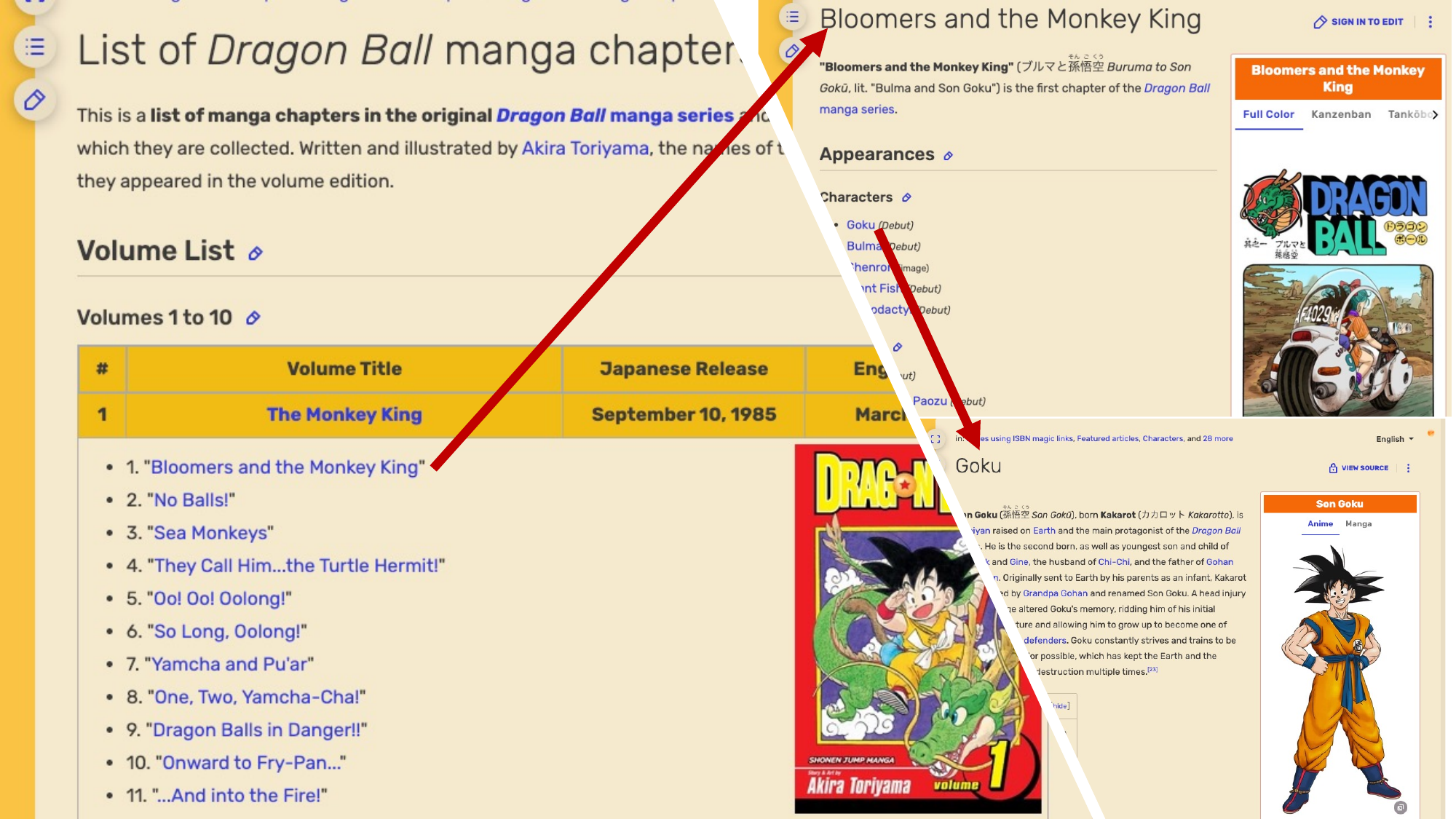}
    \caption{\textbf{Web-scraping from Fandom.} For a given series, available on Fandom, we can often scrape the list of chapters, principal characters, character appearances, and thumbnail images for the characters.
    }
    \label{fig:fandom}
\end{figure*}

Curating a character bank of principal characters for any arbitrary manga is a very tedious process. The only solution today is to read all chapters of the manga in question, manually keep track of all the characters that have been introduced and store this information in a dataset. This, of course, is a very expensive endeavour. Luckily, a lot of this heavy lifting has already been done by fans of most mangas in the PopManga dataset (see~\cref{fig:fandom}). Therefore, to compile the PopCharacters dataset, we semi-automatically scrape Fandom~\cite{fandom}, a website for fans to catalogue details regarding their favourite manga. This results in 11K+ principal characters, across 76 series, with 16K+ thumbnail images, which forms the core the PopCharacters dataset.

\subsubsection{Analysis.}

We make a few observations about the data scraped from Fandom. First, not all 84 manga series in PopManga have Fandom webpages with character information suitable for our purposes.
Second, the downloaded thumbnails for characters are often not from the manga but instead from the anime adaptation of the series, and sometimes also from the live adaptation (see~\cref{fig:popcharacters-domain-shift}). These images have a significant distribution shift in terms of the appearance of the character and are not a good representation of the manga character. Third, out of the 11K+ characters scraped, around half  do not have information on which chapters they appear in.

Given that the thumbnails scraped from Fandom are often not a good representation of the manga characters, we manually add a few `exemplar' images for each character using crops from manga chapters. However, this is too expensive to do for each of the 11k characters. For instance, in manga series like One Piece, there are more than 1.2K characters that have been catalogued by fans. While each of them might play a significant role in the story, a handful of them appear far more frequently than others (see~\cref{fig:principal-characters:one-piece}). Since the purpose of this dataset is to help name the detected characters during inference, we limit the scope of `exemplar mining' to characters that appear very frequently.

\subsubsection{Identifying frequently appearing characters.}  
\begin{figure}[!t]
    \centering
    \begin{subfigure}[b]{0.48\textwidth}
        \centering
        \includegraphics[width=\textwidth]{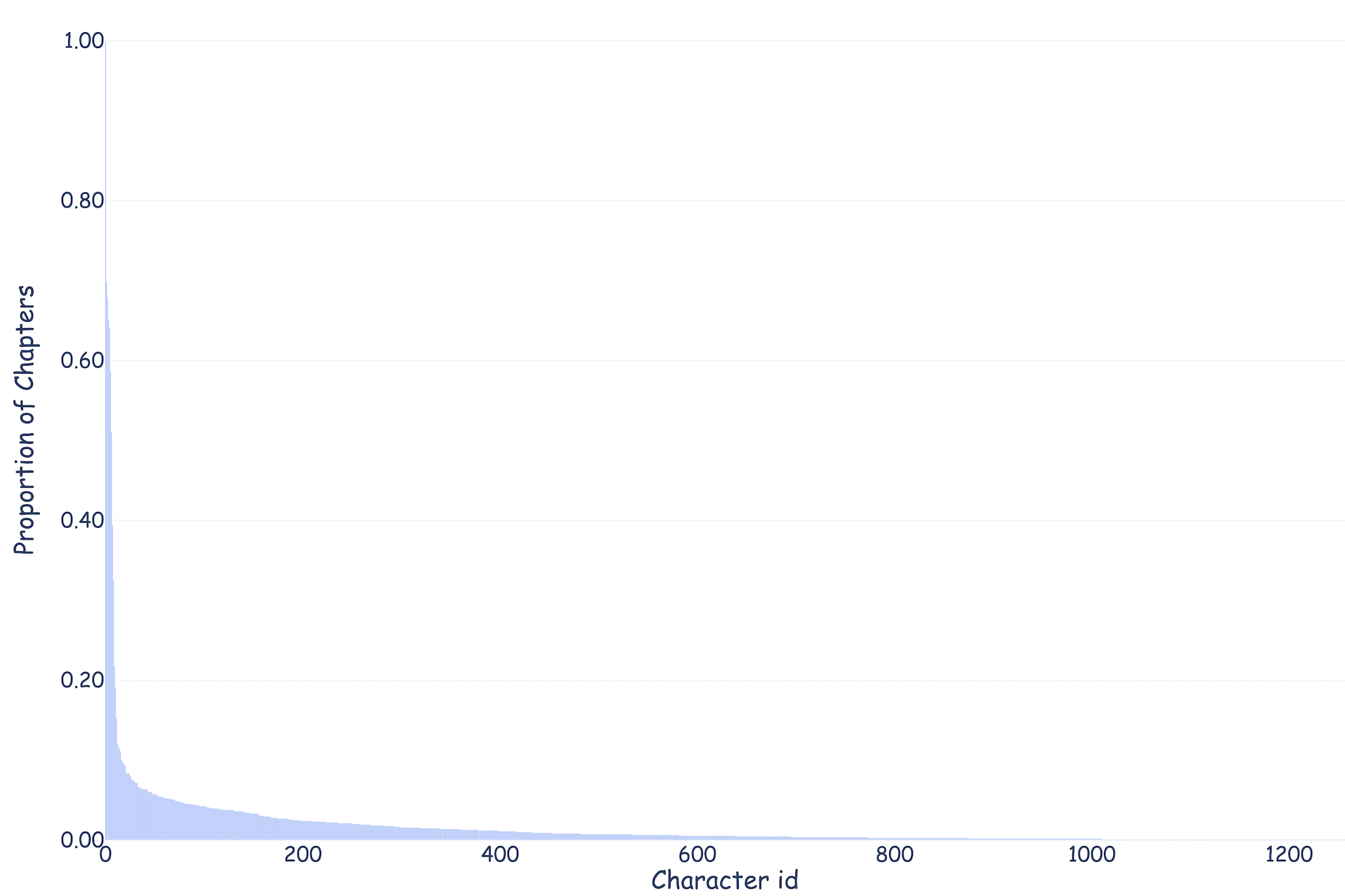}
        \caption{One Piece}
        \label{fig:principal-characters:one-piece}
    \end{subfigure}
    \hfill
    \begin{subfigure}[b]{0.48\textwidth}
        \centering
        \includegraphics[width=\textwidth]{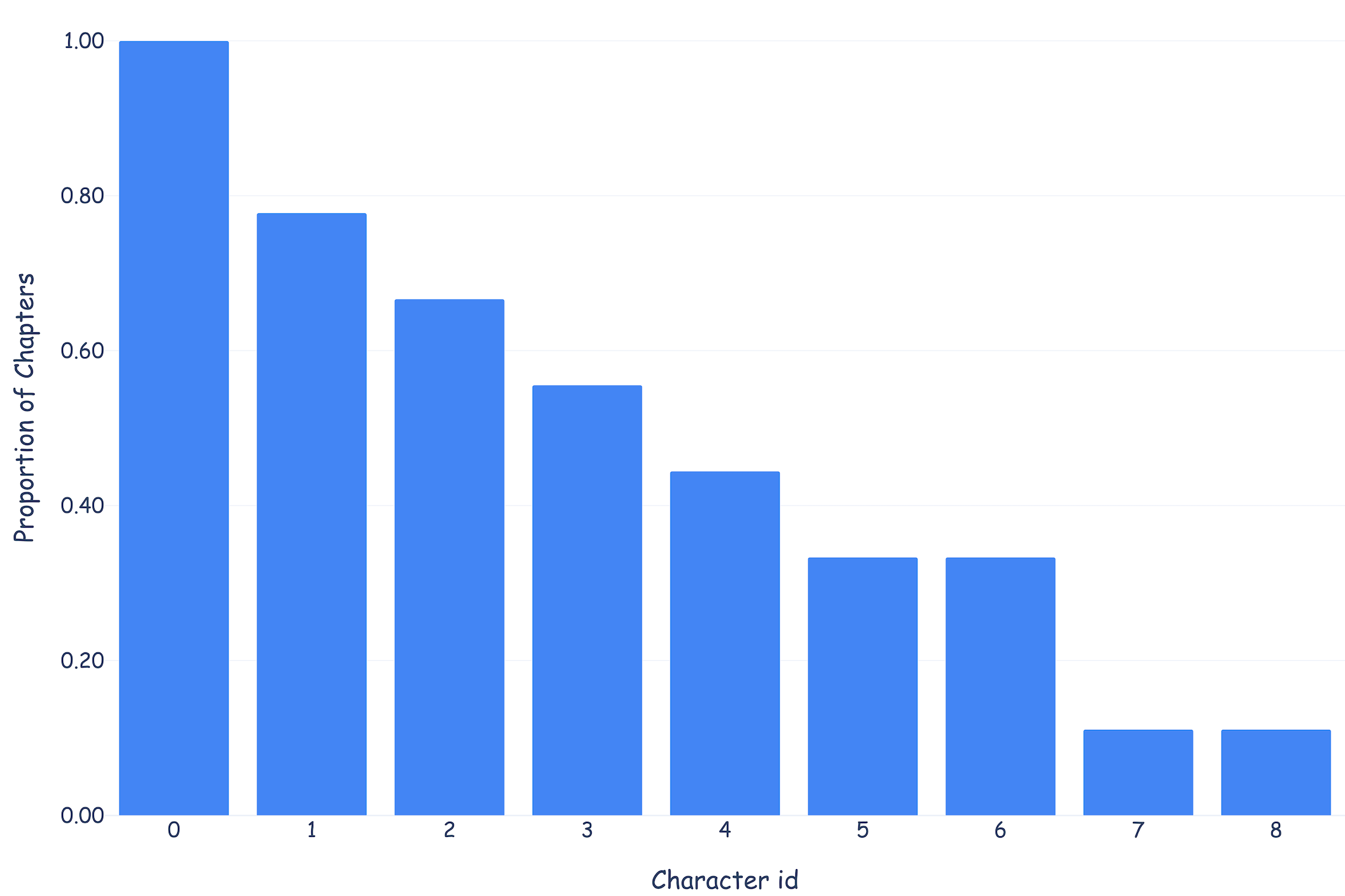}
        \caption{Vagabond}
        \label{fig:principal-characters:vagabond}
    \end{subfigure}
    \caption{Proportion of chapters for each character for two series: One Piece (a) and Vagabond (b). The proportion of chapters (y-axis) is defined as the number of chapters in which a character appears divided by the total number of chapters.
    }
    \label{fig:principal-characters}
\end{figure}

We identify characters that occur frequently using the \textit{list of chapters where a given character appears}. In particular, we consider the `character appearance frequency' (which is the proportion of chapters the character appears in, defined as the number of chapters in which a character appears divided by the total number of chapters; e.g., if a certain character appears in Chapters 1, 2, 4, and 8 when there are 16 chapters for the series, its frequency is 0.25) as a quantitative measure.
However, there is a notable challenge when using this statistic---as the number of chapters and characters are all different for different series, it is not obvious how to find a good character frequency threshold which well divides characters into the two groups and can be robustly used across the different series.
For example, as shown in~\cref{fig:principal-characters}, two series with high and low number of characters reveal stark contrast in distribution in appearance frequency across different characters.

To solve this challenge, we take the following two-step approach. First, given a series, we classify whether it has a small number of characters ($<=30$). If it does, we regard all of the characters as high-frequency characters.
Then, for each series with many characters ($>30$), we sort all the characters in decreasing order of their appearance frequency and select characters that account for up to 80\% of the entire distribution.

\begin{figure*}[h]
    \centering
    \includegraphics[width=\linewidth]{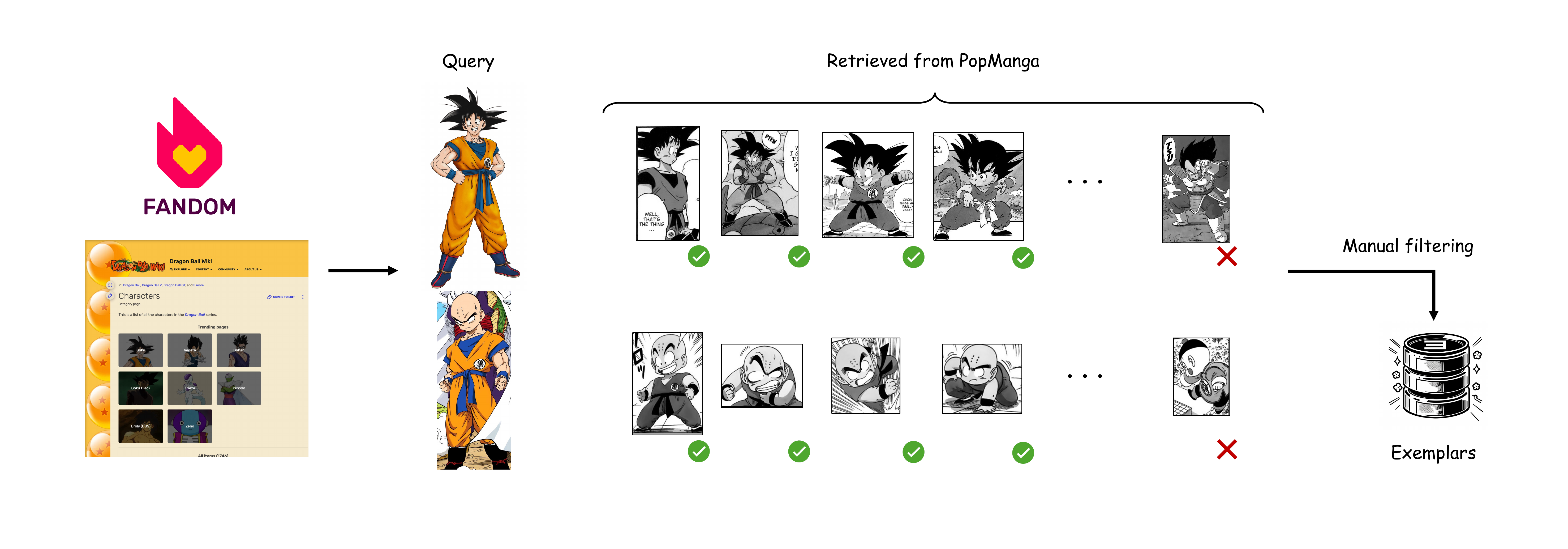}
    \caption{\textbf{Exemplar mining.} Given the thumbnail images scraped from Fandom, we retrieve matching character crops from the PopManga dataset. Manual filtering is then performed to only keep high confidence matches.
    }
    \label{fig:exemplar_mining}
\end{figure*}

\subsubsection{Exemplar mining.} 
\begin{figure}[!t]
    \centering
    \includegraphics[width=\columnwidth]{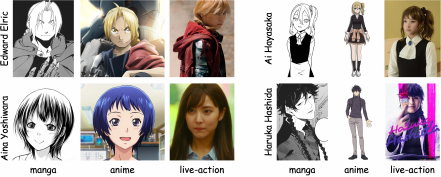}
    \caption{\textbf{Distribution of \cbd thumbnails.} Three different versions of thumbnails (\ie, manga, anime, and live-action adaptation) for four characters are shown.
    }
    \label{fig:popcharacters-domain-shift}
\end{figure}

After identifying the set of high-frequency characters, we use their scraped thumbnails as query images to retrieve high confident matches from the set of all character crops in PopManga. In the interest of diversity of appearance, we randomly select up to 20 retrieved candidate images (instead of considering top-20 most similar matches), which are then filtered manually (see~\cref{fig:exemplar_mining}). Our filtering criteria is: (i) remove false positives, (ii) remove low quality images (e.g.\ with significant occlusion by speech bubbles or other characters). In some cases there were a significant number of false positives, which does not reflect poorly on the embedding module but rather on the distribution shift of scraped thumbnails, as noted above in~\cref{fig:popcharacters-domain-shift}.

\subsubsection{Shortcomings.}
\begin{figure}[h]
    \centering
    \includegraphics[width=\columnwidth]{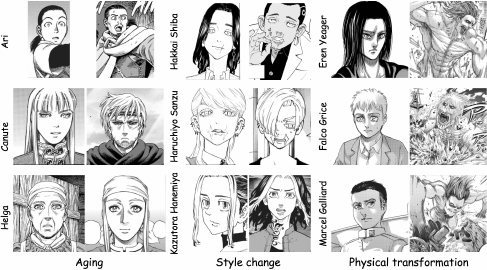}
    \caption{\textbf{Edge cases of \cbd.} Examples of characters undergoing non-negligible changes in their appearance. Typical cases of aging, style changes, and physical transformations are shown on the left, middle, and right, respectively.}
    \label{fig:popcharacters-edge-cases}
\end{figure}

Despite being the first of its kind in the research community, the PopCharacters dataset has a few shortcomings. First, unlike in movies, manga characters are much more likely to undergo radical appearance changes due to aging, magical abilities that involve transformations, or simply changes in style as shown in~\cref{fig:popcharacters-edge-cases}.
In this version of the dataset, although we include images of characters undergoing appearance changes, we do not classify differences between images of the same character, which may limit its applicability.
Second, the dataset has not been manually verified in its entirety and may contain some noise, incomplete or even incorrect information as a consequence of web-scraping. Having said that, the subset of the data that is used in evaluation (as character bank in `chapter-wide character naming') has been been through human quality assurance to ensure robust and fair benchmarking.

\subsection{\pmx}
PopManga-X is the extended version of the PopManga dataset~\cite{Sachdeva24}, wherein the test images now contain annotations for (i) speech-bubble tail bounding boxes, (ii) text-box to corresponding tail-box association, (iii) the name (identity) of each character box, and (iv) sub-classification of text boxes. In the following we provide details on the data annotation process. 

\subsubsection{Speech bubble tails.}
\begin{figure*}[h]
    \centering
    \includegraphics[width=0.6\linewidth]{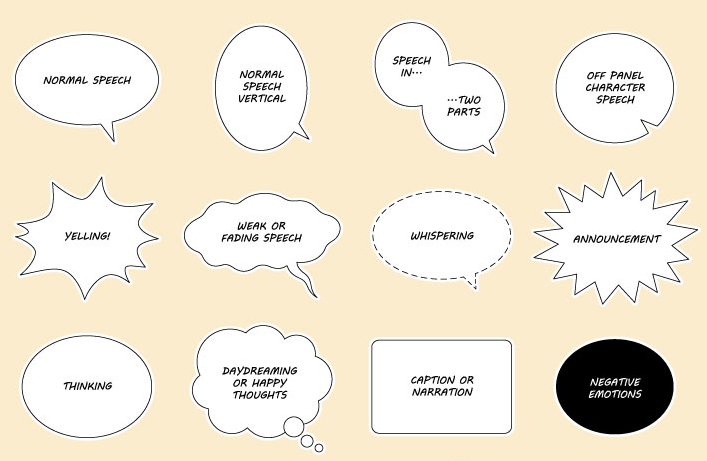}
    \caption{\textbf{Examples of speech bubbles and their intentions.} We note that this list is not complete and also not universal. Manga artists typically have their own unique conventions (e.g., it is common to also have speech bubbles with no tails as `normal speech' and not `thinking'). Image taken from animeoutline.com.
    }
    \label{fig:speech_bubbles}
\end{figure*}

In manga, speech bubbles (see~\cref{fig:speech_bubbles}) are often used to enclose dialogues, narrations etc. These speech bubbles often, not always, have tails indicating who the speaker is, or even who the speaker is not (in case of ``negative tails'' or tails pointing away from a character and towards the edge of the panel). 
We manually annotate these tail boxes by drawing tight bounding boxes around them. Additionally, we annotate the text-tail associations, which is a many-to-many relationship---for instance, a multi-part speech balloon may only have 1 tail but multiple text boxes, and, a speech bubble may have multiple tails indicating that it is being simultaneously said by multiple characters. 

\begin{figure}[h]
    \centering
    \includegraphics[width=\columnwidth]{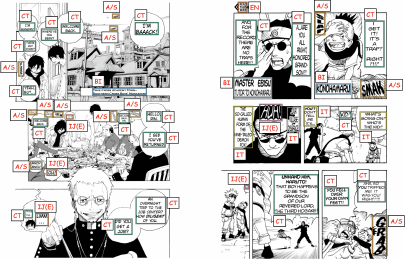}
    \caption{\textbf{Visual examples of the text categories in \pmx.} For each text box, its text category is shown in \red{red}. A/S, BI, CT, EN, IJ (E), and IT denote action/sound word, background information, conversational text, editorial note, interjection (explicit), and internal thought, respectively.
    Same text box colours within each page indicate the same text category.}
    \label{fig:text-labels}
\end{figure}

\subsubsection{Character Name Annotation.}
Previously in PopManga, character boxes had a per-page cluster ID indicating which character boxes on the page belong to the same character (\ie, have the same identity). These cluster IDs were not globally unique across the entire chapter or the series. Towards the goal of character name aware transcript generation, we label each character box with a globally unique ID (name). This is done manually by a human by considering the context of the story and using reference images from PopCharacters (where available).

\subsubsection{Text Category Annotation.}
\begin{figure}[H]
    \centering
    \includegraphics[width=\textwidth]{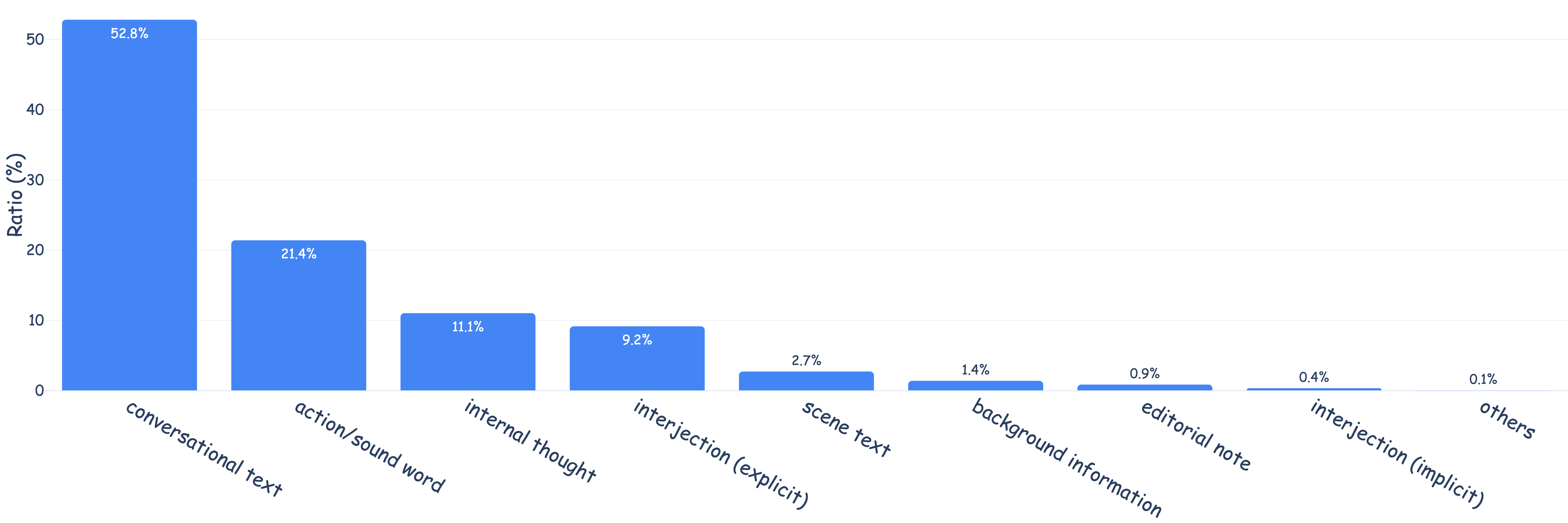}
    \caption{\textbf{Histogram of the text labels}.}
    \label{fig:text-labels-histogram}
\end{figure}

Manga pages have all sorts of texts for the reader to enjoy. However, not all of it is essential to generate a transcript and in fact can actually be a nuisance, if inappropriately included in the transcript. We manually classify the text boxes in PopManga-X to record this information. Specifically, we identified the following 9 initial text categories. \cref{fig:text-labels-histogram} shows a histogram for these text categories.
For visual examples of the text categories, see~\cref{fig:text-labels}.

\begin{itemize}
    \item Action/sound word: onomatopoeia or verbs describing action (e.g., ``bang!'' or ``Slam!'')
    \item Background information: narration or context
    \item Conversational text: conversations between characters
    \item Internal thought: texts for internal thoughts of characters
    \item Explicit interjection: interjections that are meant to be shown to other characters in the same scene
    \item Implicit interjection: interjections that are not supposed to be noticed by other characters or  interjections in an internal thought
    \item Editorial note: book/chapter names, page number, or meta-level information that is not relevant to the story
    \item Scene text: texts that are part of objects such as signs
    \item Others.
\end{itemize}

We then group these 9 categories into two: essential or non-essential for dialogue as shown in~\cref{tab:text-labels}. As a result, there are 13k+ and 7k+ texts for dialogues and non-dialogues, respectively.

{
\setlength{\tabcolsep}{3pt}
\begin{table}[!t]
\fontsize{7.25pt}{10pt}\selectfont
\centering
\caption{\textbf{Text categories.} The terms ``bkg info'' and ``conv. text'' refer to background information and conversational text, respectively.}
\vspace{-3mm}
\begin{tabular}{l|c}
& text category\\ \shline
Non-essential & action/sound word, editorial note, scene text, others\\
Essential & bkg info, conv.\ text, interjections (explicit and implicit), internal thought
\end{tabular}
\label{tab:text-labels}
\end{table}
}

\section{More on semi-supervised training}
The training datasets used to train the detection and association model is a mixed bag of unlabelled (most), partially labelled (some) and comprehensively labelled (few) images. We utilise~\cref{algo:ssl} to train our model in a semi-supervised way.

\begin{algorithm}[H]
\caption{Model Training Procedure}
\begin{algorithmic}[1]
\State \textbf{Input:} Dataset $D_l$ labeled subset, $D_u$ unlabeled subset
\State \textbf{Initialise:} Model parameters, $\theta_{init}$
\State \textbf{Phase 0: Mine \textit{partial} pseudo-labels} (no tails etc.)
\For{each $x_i$ in $D_u$}
    \State $\hat{y}_i \leftarrow \text{predict}(x_i)$ using Magi~\cite{Sachdeva24}
\EndFor
\State \textbf{Phase 1: Warm-up}
\For{each $x_i$ in $D_u$}
        \State Train model, using partial pseudo-labels $\hat{y}_i$
\EndFor
\State Update model parameters $\theta_{init}\rightarrow\theta_{interim}$
\State \textbf{Phase 2: SSL training}
\Repeat
    \State \textbf{Phase 2a: Train on labelled data}
    \For{each epoch}
        \State Train model on $D_l$
    \EndFor
    \State Update model parameters $\theta_{interim}\rightarrow\theta_{tuned}$
    \State \textbf{Phase 2b: Mine \textit{complete} pseudo-labels}
    \For{each $x_i$ in $D_u$}
        \State $\hat{y}_i \leftarrow \text{predict}(x_i)$ using model $\theta_{tuned}$
    \EndFor
    \State \textbf{Phase 2c: Re-train on pseudo labels}
    \State Initialise model with $\theta_{init}$
    \For{each epoch}
        \State Train on $D_u$ using complete pseudo labels $\hat{y}$
    \EndFor
    \State Update model parameters: $\theta_{init}\rightarrow\theta_{interim}$
\Until{fixed number of cycles}
\State \textbf{Phase 3: Fine-tuning}
\For{each epoch}
    \State Train model on $D_l$
\EndFor
\State Update model parameters $\theta_{interim}\rightarrow\theta_{tuned}$
\end{algorithmic}
\label{algo:ssl}
\end{algorithm}

\section{More on edge prediction evaluation}
The detection and association model is designed to output three kinds of edges: (i) character to character, (ii) text to character, and (iii) text to tail. In this section we provide more details regarding how our model's edge predictions are evaluated and the design decisions.

\subsubsection{Character-character edges.}
\begin{figure*}[h]
    \centering
    \includegraphics[width=\linewidth]{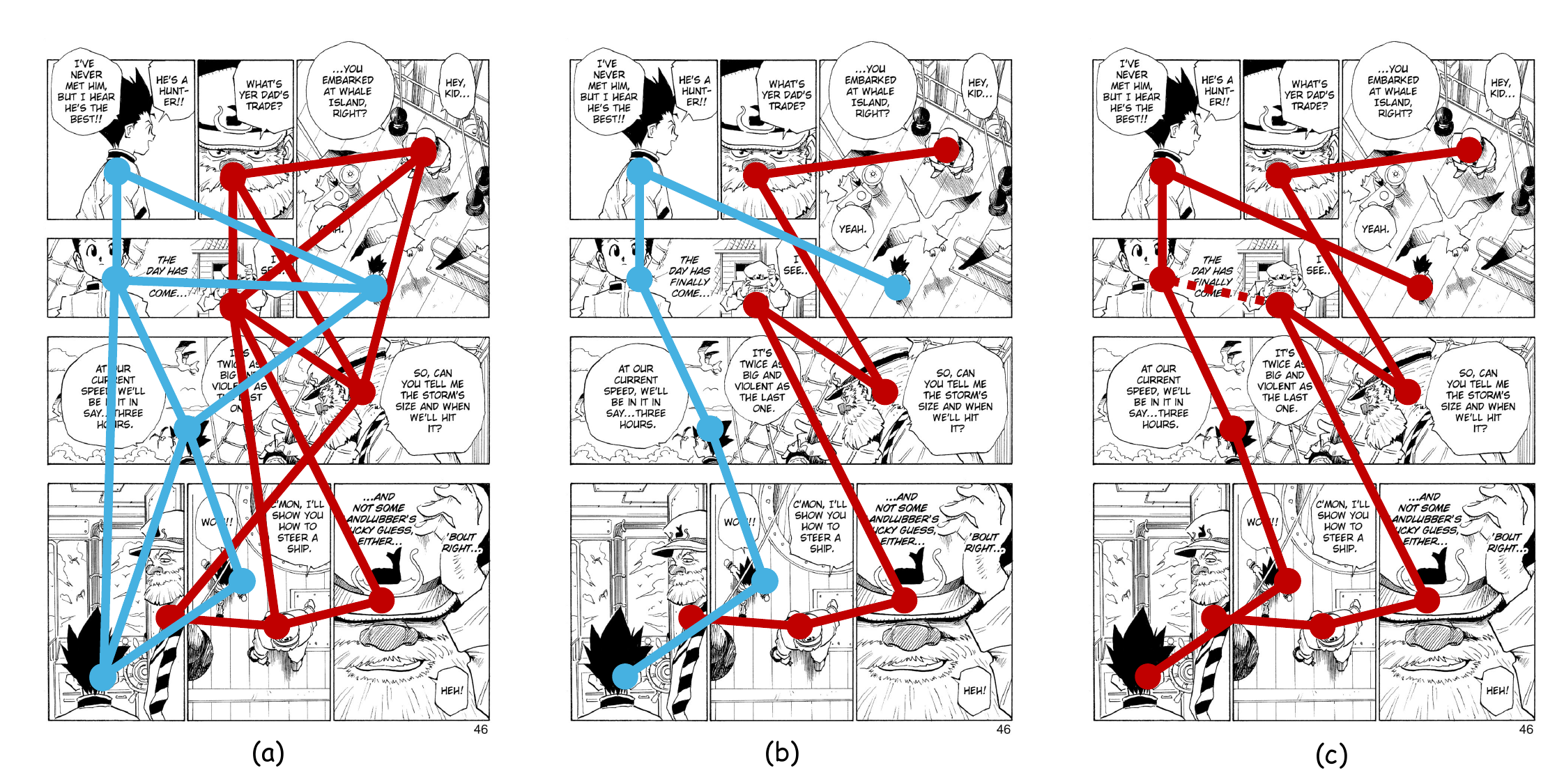}
    \caption{\textbf{Demonstrating transitive nature of character-character edges.} Even though there are significantly more number of edges in (a) than (b), they both have the same prediction for character clusters (connected components). On the contrary, (b) and (c) only differ by 1 edge (marked in dashed), yet it completely changes the predicted character clusters.
    }
    \label{fig:char-char-edges}
\end{figure*}

The important thing to note about character-character edges is that they are transitive in nature, see~\cref{fig:char-char-edges}. Therefore, the evaluation setting is that of cluster prediction and the metrics used---AMI, NMI, R-P, P@1, MAP@R, MRR---are the ones commonly used in clustering literature. Their implementation is taken from~\cite{Musgrave2020PyTorchML}. These metrics better reflect the task at hand than directly measuring the edge prediction quality, as is done below for other two types of edges.

\subsubsection{Text-tail edges.}
\begin{figure}[H]
    \centering
    \includegraphics[width=\linewidth]{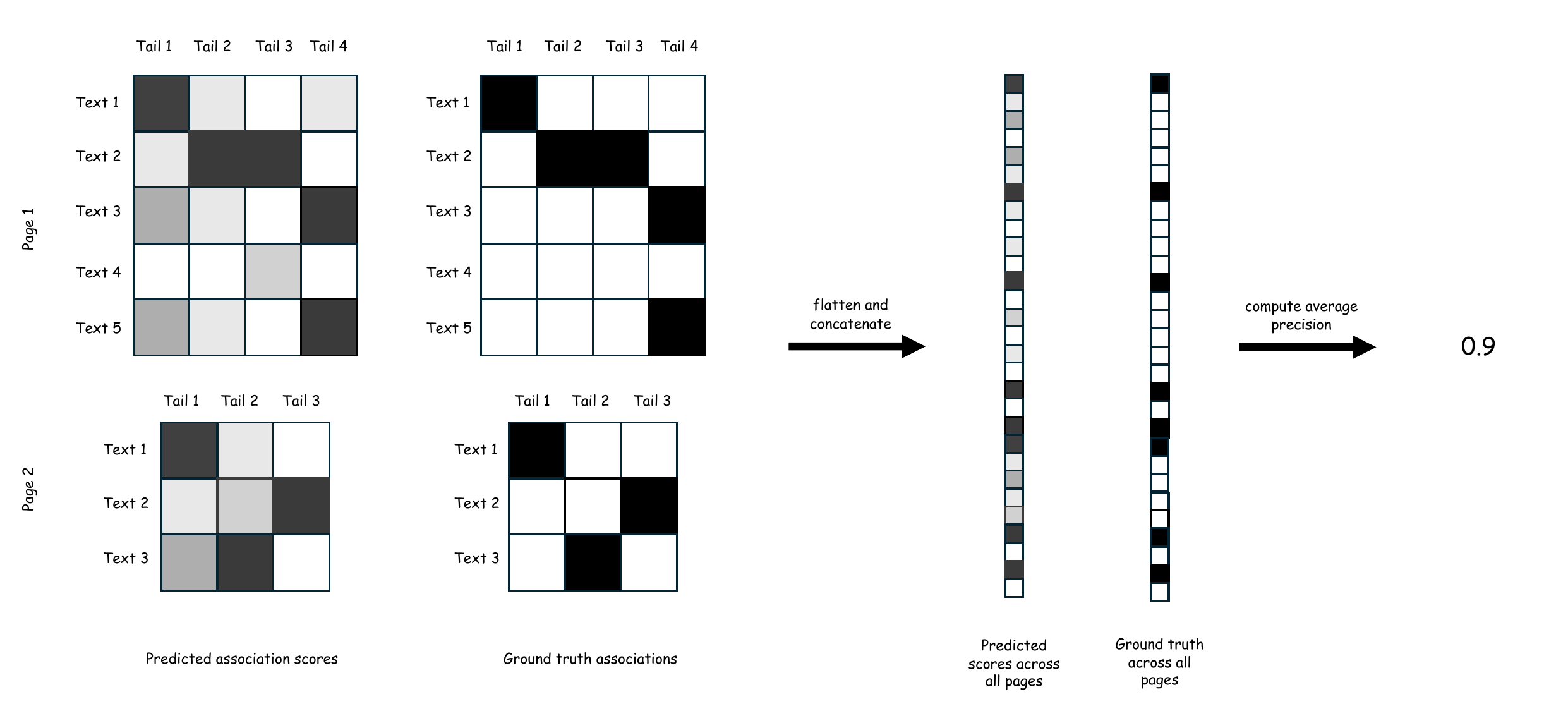}
    \caption{\textbf{Text-tail (and text-character edge) evaluation process.}
    }
    \label{fig:text-tail-edges}
\end{figure}

The text-tail edges represent the relationship between text boxes and tail boxes, \ie, whether a given tail corresponds to a given text box. This can be a many-to-many relationship, \ie, 1 or more text boxes can have a 0 or more tails. For instance, a multi-part speech bubble has 2+ text boxes which may only have a single tail box, or a single text box may have many tails indicating the case where multiple speakers simultaneously say something. To evaluate all these cases in a unified fashion, we treat it as a binary classification problem, and compute the average precision metric, as shown in~\cref{fig:text-tail-edges}.

\subsubsection{Text-character edges.}
The text-character edges embed the speaker diarisation information, \ie, which text box is said by which character. The important thing here is to note the distinction between (i) text box to character box association, and (ii) text box to character identity association.
While the model outputs the former, we actually care about the latter in terms of generating the transcript. In other words, `text box 1 is associated to character box 1' is no different than `text box 1 is associated to character box 2', \textit{if} character box 1 and 2 belong to the same character. This text to character \textit{identity} evaluation setting (instead of text to character \textit{box}) is even more important when the speaker of a text box is not present in the same panel as the text, but is present in preceding and subsequent panels.
In such cases, it is almost arbitrary as to which particular character box is the right association; however, there is only a single correct character identity that must be associated with this text.

To evaluate the text-character identity predictions, we consider the text box to character box predictions by the model, and max-pool the scores for character boxes that have the same identity. Afterwards, the evaluation process is the same as for text-tail edges, as denoted in~\cref{fig:text-tail-edges}.


\section{More on character naming evaluation}
As mentioned in the paper, the purpose of this evaluation is to measure the efficacy of the method in forming chapter-wide character clusters. When evaluating a specific chapter, we sample exactly 1 exemplar image (from PopCharacters dataset) for each character that appears in this chapter. This exemplar is used as a reference image when assigning/matching crops to this character, and it was arbitrarily chosen by a human beforehand for each character in the test set and fixed for the purposes of evaluation.

The advantage of using a single exemplar image per character is that it keeps the evaluation fair as some characters have more exemplars than others. However, this introduces another variable -- ``the choice of exemplar image'' which can significantly impact the performance results (some exemplars are better representations of the character than others). To demonstrate just how significant the gap can be, we report the character naming results in~\cref{tab:character_naming_optim} when ``optimal exemplars'' are used. Specifically, instead of arbitrarily choosing and fixing an exemplar image from PopCharacters dataset, for a given character, we consider all possible crops of this character in the chapter being evaluated (using ground truth information) and use the crop which has the highest average similarity to all the other crops of this character, as the ``optimal exemplar''. For the sake of evaluation, this has a few benefits -- (i) it eliminates the ``choice of exemplar'' variable from the evaluation process, (ii) sampling the exemplar from within the chapter increases the likelihood of the exemplar being visually representative of the character, thus reducing the effect of edge case noted in~\cref{fig:popcharacters-edge-cases}, and (iii) it makes the evaluation setting agnostic to the external character bank which is desirable for future benchmarking and comparison.

As evident from~\cref{tab:character_naming_optim}, the character naming results are significantly better when ``optimal exemplars'' are used. Of course during inference we would never have such knowledge about optimal exemplars, and therefore these results are difficult to achieve in practice. An ideal case scenario during inference is that several diverse exemplars are available for each character and their average embedding is used as the representation of the character. To keep things simple, we have not deeply investigated this.
\setlength{\tabcolsep}{4pt}
\renewcommand{\arraystretch}{1.0}

\begin{table}[H]
\caption{\textbf{Character Naming Results.} We report the accuracy results, which have an upper bound of 1.0.}
\centering
\resizebox{\columnwidth}{!}{%
\begin{tabular}{l|c|c|c|c|c}
\rowcolor[HTML]{EFEFEF} embedding model & method                                                           & notes                          & exemplars     & \multicolumn{1}{l|}{\cellcolor[HTML]{EFEFEF}\pmx (Test-S)} & \multicolumn{1}{l}{\cellcolor[HTML]{EFEFEF}\pmx (Test-U)} \\ \hline
Magi                                    & K-means~\cite{lloyd1982kmeans}                                   & nclusters$=k+1$                & random, fixed & 0.3800                                                     & 0.3993                                                    \\ \hline
Magiv2                                  & K-means~\cite{lloyd1982kmeans}                                   & nclusters$=k+1$                & random, fixed & 0.4126                                                     & 0.4221                                                    \\ \hline
Magi                                    & iForest~\cite{liu2012isolation} + K-means~\cite{lloyd1982kmeans} & nclusters$=k$                  & random, fixed & 0.4710                                                     & 0.4859                                                    \\ \hline
Magiv2                                  & iForest~\cite{liu2012isolation} + K-means~\cite{lloyd1982kmeans} & nclusters$=k$                  & random, fixed & 0.5298                                                     & 0.5096                                                    \\ \hline
Magi                                    & Constraint Optimisation (Ours)                                   & Predicted per-page constraints & random, fixed & 0.6637                                                     & 0.7058                                                    \\ \hline
Magiv2                                  & Constraint Optimisation (Ours)                                   & Predicted per-page constraints & random, fixed & 0.7273                                                     & 0.7530                                                    \\ \hline
Magi                                    & Constraint Optimisation (Ours)                                   & Predicted per-page constraints & optimal       & 0.8164                                                     & 0.8375                                                    \\ \hline
Magiv2                                  & Constraint Optimisation (Ours)                                   & Predicted per-page constraints & optimal       & \textbf{0.8735}                                            & \textbf{0.8770}                                           \\ \hline
\rowcolor[HTML]{EFEFEF} Magi            & Constraint Optimisation (Ours)                                   & GT per-page constraints        & random, fixed & 0.7445                                                     & 0.7975                                                    \\ \hline
\rowcolor[HTML]{EFEFEF} Magiv2          & Constraint Optimisation (Ours)                                   & GT per-page constraints        & random, fixed & 0.7987                                                     & 0.8786                                                    \\ \hline
\rowcolor[HTML]{EFEFEF} Magi            & Constraint Optimisation (Ours)                                   & GT per-page constraints        & optimal       & 0.8579                                                     & 0.8786                                                    \\ \hline
\rowcolor[HTML]{EFEFEF} Magiv2          & Constraint Optimisation (Ours)                                   & GT per-page constraints        & optimal       & \textbf{0.9219}                                            & \textbf{0.9302}                                          
\end{tabular}
}
\label{tab:character_naming_optim}
\end{table}

\section{More on transcript generation}
After detection (characters, texts, panels and tails) and association (character-character, text-character, text-tail), and chapter-wide character naming, generating the transcript is relatively straightforward. As mentioned in the paper, this is a four-step process: (i) filtering non-essential texts, (ii) text ordering, (iii) OCR, and (iv) generating the transcript using the predicted text-character associations and character names. The implementation for the ordering algorithm and the OCR model have been directly taken from~\cite{Sachdeva24} as the purpose of this work is not to improve on these.

Beyond that we highlight a few design decisions that can be made while generating the transcripts to make them more robust to model's mistakes and ensure narrative consistency.
First, low-confidence speaker predictions for essential texts can be rendered as `<unsure>' in the transcript, rather than confusing the reader.
Second, given that we have detected tail boxes, and matched them to their corresponding text boxes, with our method it is possible to indicate the speakers in the transcript only for the texts that have tails.
In other words, for texts without tails, it might be reasonable to just include them in the series of dialogues without indicating the speaker and let the reader infer the speakers from the context. This has two benefits: (i) it is in-line with the manga artists' intention (the fact that they chose to not draw an explicit tail for some texts), and (ii) texts without tails are usually where the model makes more mistakes.

\subsubsection{On using LLMs to enhance the transcripts.}
In this work we also investigated using LLMs to enhance the quality of the generated transcripts. While the LLMs do a very good job at fixing OCR mistakes and can be employed successfully for that purpose as a post-processing step, we were mainly interested in exploring if LLMs can leverage conversational history and context to fix speaker prediction mistakes. We discovered that text-based speaker diarisation is a very challenging problem where the ambiguity in predicting who the speaker is increases drastically as the number of speakers increases.
Often in two-person conversations, it is possible to deduce a change in speaker based on the conversation pattern; however, with three or more potential speakers, many of whom can possibly be `other', the problem becomes very challenging, and we did not have much success with using LLMs.
We also investigated training a vision-language model that leverages both vision and language cues for this task, but had limited success which we attribute to two reasons:
(i) lack of large-scale ground truth annotations (our training was largely done on pseudo-annotations mined for large-scale data which is quite noisy), and (ii) the inherent imbalance in the data (most texts in fact can be attributed to the correct speaker simply by picking the nearest one, therefore the signal for language during training is rather weak).



%
%


\end{document}